\documentclass[a4paper,fleqn]{cas-dc}

\usepackage[authoryear]{natbib}
\usepackage[utf8x]{inputenc}
\usepackage{newunicodechar}
\newunicodechar{≥}{\geq}
\newunicodechar{↔}{\leftrightarrow{}}
\usepackage{amsmath,amsfonts}
\usepackage{booktabs}
\usepackage{multirow}
\usepackage{array}
\usepackage{rotating}
\usepackage{threeparttable} % For table notes
\usepackage{graphicx}      % For resizebox
\usepackage{algorithmic}
\usepackage{algorithm}
\usepackage{array}
\usepackage{xcolor}
\usepackage{bm}
\usepackage{algorithmic}
\usepackage{hyperref}
\usepackage[caption=false,font=normalsize,labelfont=sf,textfont=sf]{subfig}
\usepackage{textcomp}
\usepackage{booktabs}
\usepackage{multicol}

\usepackage{multirow}
\usepackage{stfloats}
\usepackage{adjustbox}
\usepackage{float}
\usepackage{url}
\usepackage{verbatim}
\usepackage{colortbl}

\usepackage{graphicx}

\def\tsc#1{\csdef{#1}{\textsc{\lowercase{#1}}\xspace}}
\tsc{WGM}
\tsc{QE}
\begin{document}
\let\WriteBookmarks\relax
\def\floatpagepagefraction{1}
\def\textpagefraction{.001}

\shorttitle{}    

\shortauthors{}  

\title [mode = title]{A Road-Conditioned Traffic Movie Prediction Network with Spatiotemporal and Structure-Consistent Learning}  

\tnotemark[1] 

\tnotetext[1]{} 



\author[1]{Joshua Kofi Asamoah}[orcid=0009-0002-3258-0479]

% \cormark[1]

\fnmark[1]

\ead{joshua.asamoah@ndsu.edu}

\ead[url]{https://joshuakasamoah.github.io/}

\credit{Conceptualization of this study, Methodology, Writing - Original draft preparation, Writing - Review}

\affiliation[1]{organization={North Dakota State University},
            addressline={1410 14th Avenue Offerdahl North Building, CIE 201}, 
            city={Fargo},
%          citysep={}, % Uncomment if no comma needed between city and postcode
            postcode={58102}, 
            state={North Dakota},
            country={United States}}

\author[1]{Blessing Agyei Kyem}[orcid=0009-0006-6360-6386]

% Footnote of the second author
\fnmark[2]

% Email id of the second author
\ead{blessing.agyeikyem@ndsu.edu}

\ead[url]{https://blessing-agyei-kyem.github.io/}

\credit{Data Curation, Methodology, Writing - Review and Editing, Validation}

% \affiliation[1]{organization={North Dakota State University},
%             city={Fargo},
% %          citysep={}, % Uncomment if no comma needed between city and postcode
%             postcode={58102}, 
%             state={North Dakota},
%             country={United States}}

\author[1]{Armstrong Aboah}[orcid=0000-0002-1605-1545]
\cormark[1]
\ead{armstrong.aboah@ndsu.edu}
\ead[url]{https://aboaharmstrong.vercel.app/}
\credit{Conceptualization of this study, Methodology, Writing - Original draft preparation, Writing - Review and Editing, Supervision}

\cortext[cor1]{Corresponding Author}

\fntext[1]{}

\nonumnote{}

\begin{abstract}
City-wide traffic forecasting is essential for congestion management, route guidance, and intelligent transportation systems, but reliable prediction remains difficult when traffic must be forecast as future spatial maps over an entire urban network. Recent traffic movie prediction methods have improved frame-level forecasting accuracy, yet many still treat the task mainly as image reconstruction. As a result, they may generate future traffic maps that are numerically close to the ground truth but weakly constrained by road layout, connectivity, travel direction, and congestion propagation. This limitation becomes more critical under cross-city settings, where both traffic behavior and road structure change substantially across urban environments. To address this problem, this study proposes RCSNet, a road-conditioned spatiotemporal network that reformulates traffic movie prediction as topology-guided future-state generation. RCSNet extracts topology-aware road representations from static road maps, captures multi-horizon traffic dynamics from historical observations, aligns directional traffic features with local road structure, and progressively generates future traffic maps to improve temporal consistency. A structure-consistent learning objective further encourages predictions to remain accurate, road-aligned, and spatially stable over the forecasting horizon. Experiments on multiple cities show that RCSNet consistently improves both forecasting accuracy and structural consistency. In same-city forecasting on Berlin, Antwerp, and Moscow, RCSNet reduces average MAE, MSE, and RMSE by 11.5\%, 10.0\%, and 5.1\%, respectively, compared with the closest baseline. In cross-city testing on unseen Chicago and Bangkok, it reduces RMSE by 10.6\% and 10.5\% without target-city fine-tuning. Additional qualitative, horizon-wise, road-structure, explainability, statistical, and efficiency analyses demonstrate that RCSNet produces more accurate, transferable, road-aligned, and computationally efficient traffic forecasts. 

\end{abstract}

% \includegraphics[width=0.8\linewidth]{rcsnet_overall.pdf}

\begin{highlights}
\item A road-conditioned traffic movie prediction framework is developed using static road topology as structural guidance.

\item Topology-aware road features are extracted from road occupancy, orientation, connectivity, and intersection-related cues.

\item Multi-horizon temporal encoding captures short-, medium-, and long-range traffic dynamics from historical traffic maps.

\item Direction-aware road-traffic fusion aligns dynamic traffic features with local road structure and movement direction.

\item Scales toward statewide real-time traffic estimation to support road safety, congestion management, and transportation intelligence.
\end{highlights}

%\nocite{*}

% Keywords

\maketitle

% Main text
\section{Introduction}\label{introduction}
With the rapid growth of modern cities, urban road networks have become larger, busier, and more complex \cite{asamoah2025saam}. This expansion has increased the need for reliable traffic management systems that can reduce congestion, improve road safety, and support more efficient mobility \cite{meena2020traffic,pravin2025review,asamoah2026querymotion}. Traffic prediction is central to these systems because it provides advance knowledge of future traffic conditions \cite{wu2018hybrid,shoman2024graph,li2023dynamic}. This knowledge helps transportation agencies improve signal control, manage congestion, support route guidance, and respond more effectively to changing traffic patterns \cite{boukerche2020artificial,chavhan2020prediction}. For this reason, traffic prediction has become an important research area in intelligent transportation systems \cite{,pritha2024smart}.

Within this research area, earlier studies mainly used statistical methods such as ARIMA models \cite{ho1998use,newbold1983arima,tseng2002fuzzy} and Kalman filtering \cite{morrison1977kalman,sharma2020blind}. However, they struggled to capture nonlinear spatiotemporal dependencies in urban networks, where congestion can spread across connected road segments and vary with time, demand, and road structure \cite{jing2025highway,zhu2025deep,duan2018trade}. To overcome these limitations, recent studies have increasingly adopted deep learning models for traffic prediction \cite{shoman2024graph,agyei2026pavecap}. Convolutional neural networks have been used to extract spatial patterns from grid-based traffic maps \cite{han2021dynamic, yu2017spatio, salehi2024experimental}, while recurrent models \cite{wu2016short,ma2015long,fu2016using} and temporal convolutional networks \cite{yu2017spatio} have been used to capture traffic evolution over time. Graph neural networks \cite{wu2019graph,choi2022graph,hermes2022graph} further extended this direction by representing road networks as graphs and modeling dependencies between connected road segments. These methods have improved traffic forecasting performance by learning richer spatial and temporal representations from data. 

Building on this progress, recent research has shifted toward city-wide traffic map forecasting, where traffic conditions are represented as sequences of spatial grids rather than isolated sensor measurements \cite{eichenberger2022traffic4cast,neun2023traffic4cast,martin2019traffic4cast}. This setting is commonly formulated as traffic movie prediction, in which historical traffic frames are used to generate future frames over a city-scale road network \cite{kopp2021traffic4cast}. The formulation is useful because it captures traffic evolution over broad urban areas and supports prediction in locations where fixed sensor coverage is limited. Despite its advantages, traffic movie prediction remains challenging when models are applied to cities that were not seen during training. Cross-city transfer changes both the observed traffic behavior and the road structure that shapes it. Traffic demand, density, and daily movement patterns vary across cities, while road geometry, intersections, corridor continuity, and travel directions also differ. As a result, a model that learns strong frame-to-frame relationships in one city may still struggle in another if those relationships are not tied to the target city's road structure.

Existing methods address this challenge from different angles, but they still leave important gaps. Encoder-decoder and U-Net based methods \cite{choi2020utilizing,santokhi2021dual} provide strong frame reconstruction, yet they usually treat traffic maps as image sequences, which makes it difficult to enforce road-topology constraints. Graph-based models \cite{hermes2022graph} introduce connectivity information and move closer to the physical road network, but they often do not fully link topology with directional traffic flow and multi-step evolution. Transformer-based models \cite{lu2021learning,wiedemann2021traffic,kyem2025context} capture long-range dependencies through attention, but attention alone does not guarantee that future traffic remains aligned with valid road regions. Transfer-oriented and temporal adaptation methods \cite{konyakhin2021solving} improve cross-city robustness, but they mainly focus on feature transfer rather than using the target road network as an explicit guide for future traffic generation. These model-level limitations become more serious as the prediction horizon increases. Traffic does not evolve at one fixed temporal scale. Short-term fluctuations, medium-term congestion propagation, and longer-term movement trends can occur within the same forecasting window. However, many models compress the input sequence into a single temporal representation, which weakens their ability to separate these different dynamics. In addition, future frames are often generated with limited step-by-step refinement, so errors from earlier horizons can affect later predictions. Standard pixel-level losses further compound this problem because they penalize average numerical error but do not explicitly discourage traffic from spreading into non-road regions or becoming spatially diffuse.

To address these limitations, this study proposes RCSNet, a road-conditioned spatiotemporal framework that reformulates traffic movie prediction as topology-guided future-state generation. Instead of predicting future traffic maps only from historical traffic frames, RCSNet uses the static road network as a structural prior that guides where traffic can occur, how directional flows should move, and how future states should evolve over time. The framework first extracts topology-aware road features from the static map and learns dynamic traffic representations from the input sequence through multi-horizon temporal encoding. These two sources are then combined through direction-aware road-traffic fusion so that traffic dynamics are interpreted in relation to the local road structure. Future traffic maps are generated progressively across the prediction horizon, allowing each step to build on the evolving forecast context rather than producing all future frames as independent outputs.  The learning objective is also designed to support this formulation. In addition to the standard prediction loss between the generated and ground-truth traffic maps, RCSNet uses a structure-consistent loss that encourages the predicted traffic distribution to remain aligned with valid road regions and temporally coherent across future frames. This loss formulation helps reduce spatially diffused predictions, unrealistic off-road activations, and unstable long-horizon forecasts. As a result, RCSNet moves beyond simple frame reconstruction toward physically grounded traffic forecasting, producing predictions that are accurate in value, consistent with the road network, stable across time, and more transferable to unseen cities.

To this end, the contributions of this paper are as follows:

\begin{enumerate}
    \item We develop RCSNet, a road-conditioned spatiotemporal framework that reformulates traffic movie prediction as topology-guided future-state generation rather than conventional frame reconstruction.

    \item We develop an integrated road-traffic learning design that combines topology-aware road representation, multi-horizon temporal encoding, and direction-aware fusion to jointly model road structure, traffic dynamics, and movement direction.

     \item We formulate a progressive structure-consistent prediction strategy that generates future traffic maps sequentially while using a combined learning objective to improve prediction accuracy, road alignment, and temporal coherence.

    \item We validate the proposed design through extensive same-city and cross-city experiments across multiple urban environments, where RCSNet consistently improves forecasting accuracy, transferability, road-structure consistency and computational efficiency over representative baselines.
\end{enumerate}

The remainder of this paper is organized as follows. Section~\ref{related} reviews related studies on traffic forecasting and traffic movie prediction, with emphasis on city-wide spatiotemporal modeling, road-structure-aware learning, and cross-city generalization. Section~\ref{methodology} presents the proposed RCSNet framework, including topology-aware road representation, multi-horizon temporal encoding, direction-aware road-traffic fusion, progressive multi-step decoding, and structure-consistent learning. Section~\ref{experiment} describes the datasets, evaluation metrics, baseline models, and implementation settings. Section~\ref{results} reports the same-city and cross-city results, followed by ablation analysis, forecast horizon analysis, qualitative visualization, road-structure consistency analysis, model explainability, statistical evaluation, and computational efficiency assessment. Section~\ref{conclusion} concludes the paper with the main findings and future research directions.

\section{Related Studies}
\label{related}
This section reviews the most relevant studies related to city-scale traffic forecasting, road-structure-aware prediction, and multi-step spatiotemporal modeling. The discussion highlights the main advances in these directions and identifies the limitations that motivate the proposed framework.

\subsection{Deep Learning for City-Scale Traffic Forecasting}
Early traffic forecasting studies mainly relied on statistical and shallow machine learning models, including Kalman filtering \cite{chen2015forecasting}, ARIMA-type methods \cite{ding2011forecasting,kumar2015short,williams1999modeling}, and support vector regression \cite{hong2006highway,lin2022using,liu2018short}. These approaches were effective for modeling short-term temporal patterns and laid the foundation for later traffic prediction research. However, their performance often degraded in large urban networks, where traffic behavior is highly nonlinear, spatially heterogeneous, and strongly interconnected \cite{miglani2019deep,denteh2025integrating}.

This limitation led to the adoption of deep learning methods, which offered a more direct way to learn spatiotemporal traffic dependencies from data \cite{ahn2016highway, zhang2013improved, mai2014short}. For instance, Zhang et al. introduced ST-ResNet \cite{zhang2017deep}, which modeled temporal closeness, period, and trend through residual convolutional branches and demonstrated that grid-based traffic prediction could be learned effectively with deep spatial feature extraction. Yao et al. later proposed DMVST-Net \cite{yao2018deep}, which extended this direction by combining local convolutional features \cite{han2021dynamic, salehi2024experimental}, temporal dependency modeling, and semantic similarity between regions for taxi demand prediction. As the field matured, researchers also began to model traffic directly on transportation networks rather than only on spatial grids. Li et al. proposed DCRNN \cite{li2017diffusion}, which used diffusion convolution and an encoder-decoder recurrent structure to capture spatial dependency on directed graphs and temporal dependency across successive observations. This line of work was strengthened further by Graph WaveNet \cite{wu2019graph}, AGCRN \cite{bai2020adaptive}, and DMSTGCN \cite{han2021dynamic}, which introduced adaptive graph learning together with dilated temporal convolutions to better capture hidden spatial relations and longer temporal contexts. 

Subsequent studies like DGCRN \cite{li2023dynamic} and D$^2$STGNN \cite{shao2022decoupled} extended this direction by introducing more flexible graph and attention mechanisms for traffic forecasting \cite{zhang2016dnn}. Guo et al. proposed ASTGCN \cite{guo2019attention}, which combined spatial-temporal attention with graph convolution and explicitly modeled recent, daily, and weekly traffic dependencies. Zheng et al. later introduced GMAN \cite{zheng2020gman}, where an encoder-decoder architecture with multi-attention blocks and transformer attention was used to capture long-range spatiotemporal interactions more effectively. In parallel, Song et al. proposed STSGCN \cite{song2020spatial}  to model localized spatial-temporal subgraphs in a synchronous manner, while Li and Zhu developed STFGNN \cite{li2021spatial} to strengthen the joint learning of hidden spatial relations and temporal dynamics through graph fusion. These studies improved traffic forecasting by moving beyond fixed graph construction and simple temporal stacking, while also showing that stronger forecasting performance depends on effectively coupling spatial structure with temporal dynamics in a unified architecture \cite{diao2019dynamic}.

Despite these advances, spatiotemporal modeling alone does not guarantee generalization, since traffic patterns vary across cities and time periods, motivating cross-city and transfer-oriented forecasting methods that improve robustness under spatial and temporal distribution shifts.

\subsection{Cross-City Generalization and Domain Shift in Traffic Forecasting}

Recent studies have addressed traffic forecasting across cities more directly by treating it as a problem of cross-city generalization under spatial and temporal shift. One of the earlier works in this direction is the multi-task learning framework of Lu et al. \cite{lu2021learning}, where forecasting is learned jointly across multiple cities so that shared representations can improve transferability while still preserving city-specific characteristics. Choi et al. \cite{choi2020utilizing} also followed a cross-city setting through a U-Net-based solution trained across multiple urban environments, showing that shared multi-city training can improve robustness in large-scale traffic-map prediction. Konyakhin et al. \cite{konyakhin2021solving} further strengthened this line of work by combining a U-Net backbone with temporal domain adaptation, showing that cross-city robustness also depends on reducing temporal mismatch between training and testing periods. In a more explicit adaptation setting, Fang et al. \cite{fang2022transfer} proposed STAN, a spatio-temporal adaptation network for cross-city urban flow prediction, and showed that effective transfer must account for both spatial heterogeneity and temporal inconsistency across cities.

Later studies extended this direction by introducing stronger alignment and representation learning strategies for cross-city forecasting. Chen et al. \cite{chen2024semantic} proposed a semantic-fused multi-granularity transfer framework that combines semantic fusion, hierarchical clustering, graph construction, and adversarial learning to better align source and target cities. Wang et al. \cite{wang2018cross} also studied cross-city traffic prediction under data scarcity by introducing a domain fusion mechanism that incorporates auxiliary urban information into the forecasting process. In a more challenging setting, Prabowo et al. \cite{prabowo2023traffic} considered forecasting on unseen roads and showed that spatial contrastive pre-training can improve generalization when the target roads are not observed during training. At the same time, several large-scale forecasting models \cite{wiedemann2021traffic,wang2021traffic4cast,hermes2022graph,bojesomo2022swinunet3d,santokhi2021dual}, further reinforced this direction by improving robustness through stronger spatiotemporal learning, graph reasoning, attention, and domain-aware representation design. Together, these studies show that cross-city forecasting has gradually evolved from shared learning toward more deliberate cross-domain alignment, adaptation, and transferable representation learning.

\subsection{Limitations of Existing Works}

Although existing studies have improved traffic forecasting under cross-city settings, several limitations remain. First, many methods improve transferability through shared learning, domain alignment, graph modeling, or attention-based representation learning, but they do not explicitly use the target city road topology to guide how future traffic should evolve. As a result, a model may learn transferable traffic features, but still produce forecasts that are weakly aligned with the road layout, movement direction, and valid traffic regions of an unseen city. Second, most traffic movie prediction models still treat future map generation mainly as a frame reconstruction problem. This makes it difficult to distinguish between traffic values that are numerically close and predictions that are physically meaningful. For example, a forecast may achieve a low pixel-level error while still spreading traffic into non-road regions or failing to preserve corridor-level movement patterns. This limitation becomes more important in city-wide prediction, where road geometry, intersections, and directional flow strongly influence how traffic propagates. Third, existing models often compress temporal information into a single representation or generate future frames with limited step-by-step refinement. However, traffic evolves at multiple temporal scales, including short-term fluctuations, intermediate congestion propagation, and longer-term movement trends. Without explicitly modeling these different horizons, forecasting errors can grow as the prediction horizon increases, especially when the model is transferred to cities with different traffic dynamics.

These limitations show the need for a forecasting framework that jointly considers road structure, temporal evolution, directional movement, and stable multi-step generation. The proposed RCSNet addresses this need by combining topology-aware road representation, multi-horizon temporal encoding, direction-aware road-traffic fusion, progressive decoding, and structure-consistent learning within a unified road-conditioned prediction framework.

\section{Methodology}
\label{methodology}

This section presents the proposed RCSNet framework for road-conditioned traffic movie prediction. The goal is to predict future city-wide traffic maps by jointly modeling historical traffic dynamics and the static road structure that constrains traffic movement. The methodology first defines the forecasting problem and input representation, then describes the overall architecture and its four main components: topology-aware road representation, multi-horizon temporal traffic encoding, direction-aware road-traffic fusion, and progressive multi-step decoding. Finally, the structure-consistent learning objective is introduced to guide the model toward forecasts that are accurate, temporally coherent, and aligned with the physical road network.

\subsection{Problem Formulation}

Given a historical traffic sequence \(X = \{x_1, x_2, x_3, \ldots, x_{T_{in}}\}\) and a static road map \(M\), the goal is to learn a forecasting function that predicts the future traffic sequence \(Y = \{y_1, y_2, y_3, \ldots, y_{T_{out}}\}\). Each traffic state is represented as a multi-channel spatial grid describing directional traffic conditions, while the static road map provides structural information about the road layout, connectivity, and valid traffic regions.

Unlike other approaches that treat the spatial domain as a uniform grid, the proposed framework explicitly incorporates road topology as a structural prior to guide feature learning and prediction. In addition, instead of predicting all future frames independently, the model progressively refines future traffic states across the forecasting horizon to better capture temporal dependency among successive predictions.

Formally, the forecasting task is defined as
\begin{equation}
\hat{Y} = f(X, M; \Theta),
\end{equation}
where \(f(\cdot)\) denotes the proposed topology-aware road-conditioned forecasting network, \(\Theta\) represents the learnable parameters, and \(\hat{Y}\) is the predicted future traffic sequence. The objective is to estimate \(\hat{Y}\) such that it remains temporally accurate, spatially consistent, and structurally aligned with the underlying road network.

\subsection{Input Representation}

Each training sample consists of three components: a historical traffic sequence, a future target sequence, and a static road map. The historical traffic input is represented as
$\mathbf{X} \in \mathbb{R}^{B \times C \times T_{in} \times H \times W}$, where \(T_{in}\) is the number of observed frames, \(C\) is the number of traffic channels, and \(H \times W\) is the spatial grid size. The prediction target is represented as
$\mathbf{Y} \in \mathbb{R}^{B \times C \times T_{out} \times H \times W}$, where \(T_{out}\) is the number of future frames to be predicted. In this study, \(T_{in}=12\), \(T_{out}=12\), and \(C=8\).

Each traffic frame contains directional volume and speed maps. Specifically, the eight channels represent volume and speed information across multiple movement directions, allowing the model to learn both traffic intensity and motion state over the spatial grid. This representation is important for traffic movie prediction because future traffic conditions depend not only on where traffic is present, but also on how movement evolves across directions. In addition to the dynamic traffic sequence, each sample includes a normalized static road map
\(\mathbf{R} \in \mathbb{R}^{1 \times H \times W}\). This map provides the physical road layout of the city and serves as the structural input to the proposed framework.

\begin{figure*}
    \centering
    \includegraphics[width=0.8\linewidth]{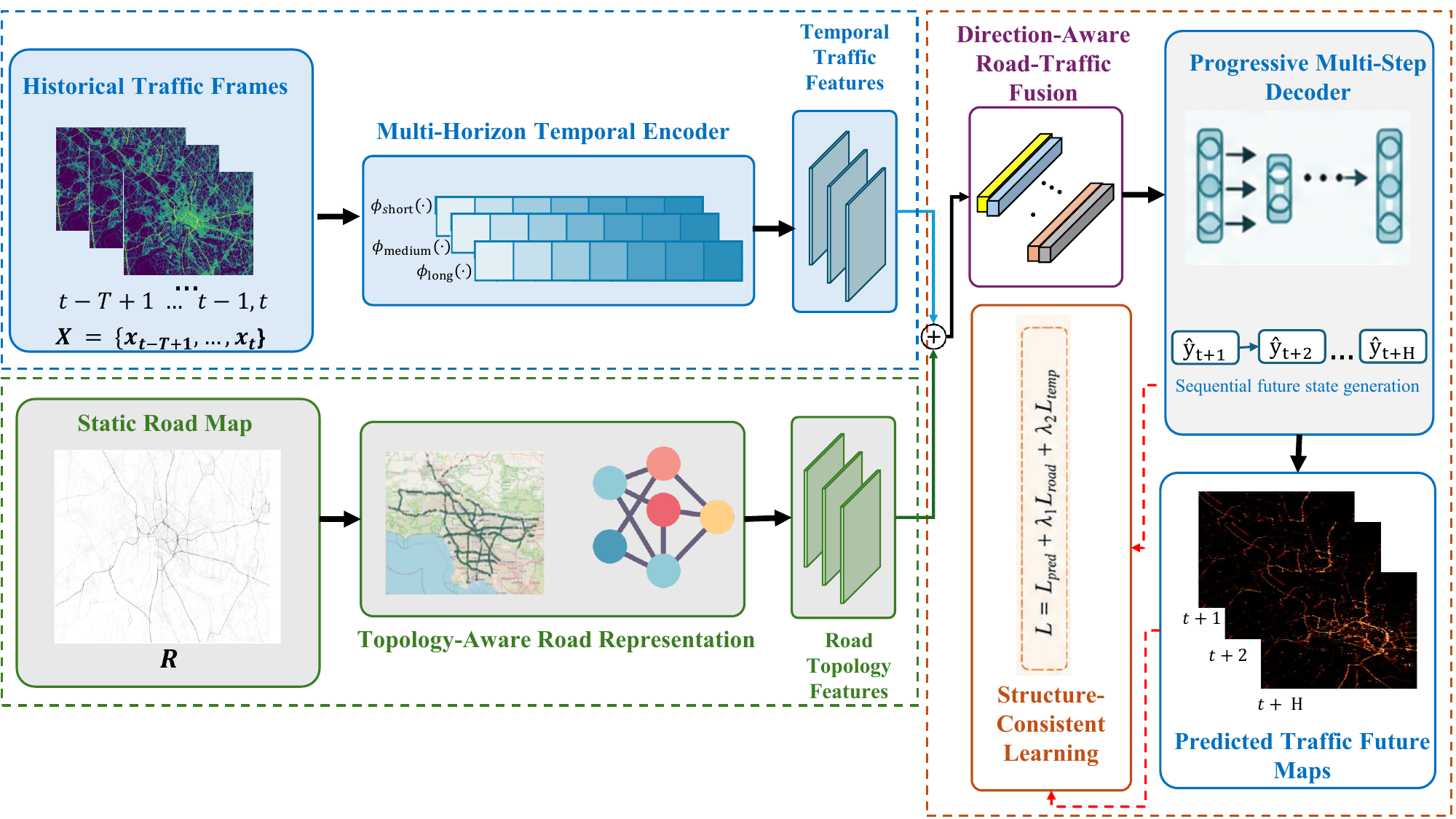}
    \caption{Overall architecture of the proposed RCSNet framework. Historical traffic frames and the static road map are processed through separate temporal and topology-aware branches, fused through direction-aware road-traffic interaction, and decoded progressively to generate future traffic maps under a structure-consistent learning objective.}    
    \label{fig:architecture}
\end{figure*}

\subsection{Proposed Architecture}

Building on the input representation, the proposed RCSNet framework forecasts future traffic states by jointly modeling dynamic traffic evolution and static road topology, as shown in Fig.~\ref{fig:architecture}. The framework consists of four main components: a topology-aware road representation module, a multi-horizon temporal traffic encoder, a direction-aware road-traffic fusion module, and a progressive multi-step forecast decoder. The following subsections describe how each component contributes to road-conditioned traffic movie prediction.

\vspace{0.8em}
\subsubsection{Topology-Aware Road Representation Module}

\begin{figure}
    \centering
    \includegraphics[width=1\linewidth, trim=0 2cm 0 2cm, clip]{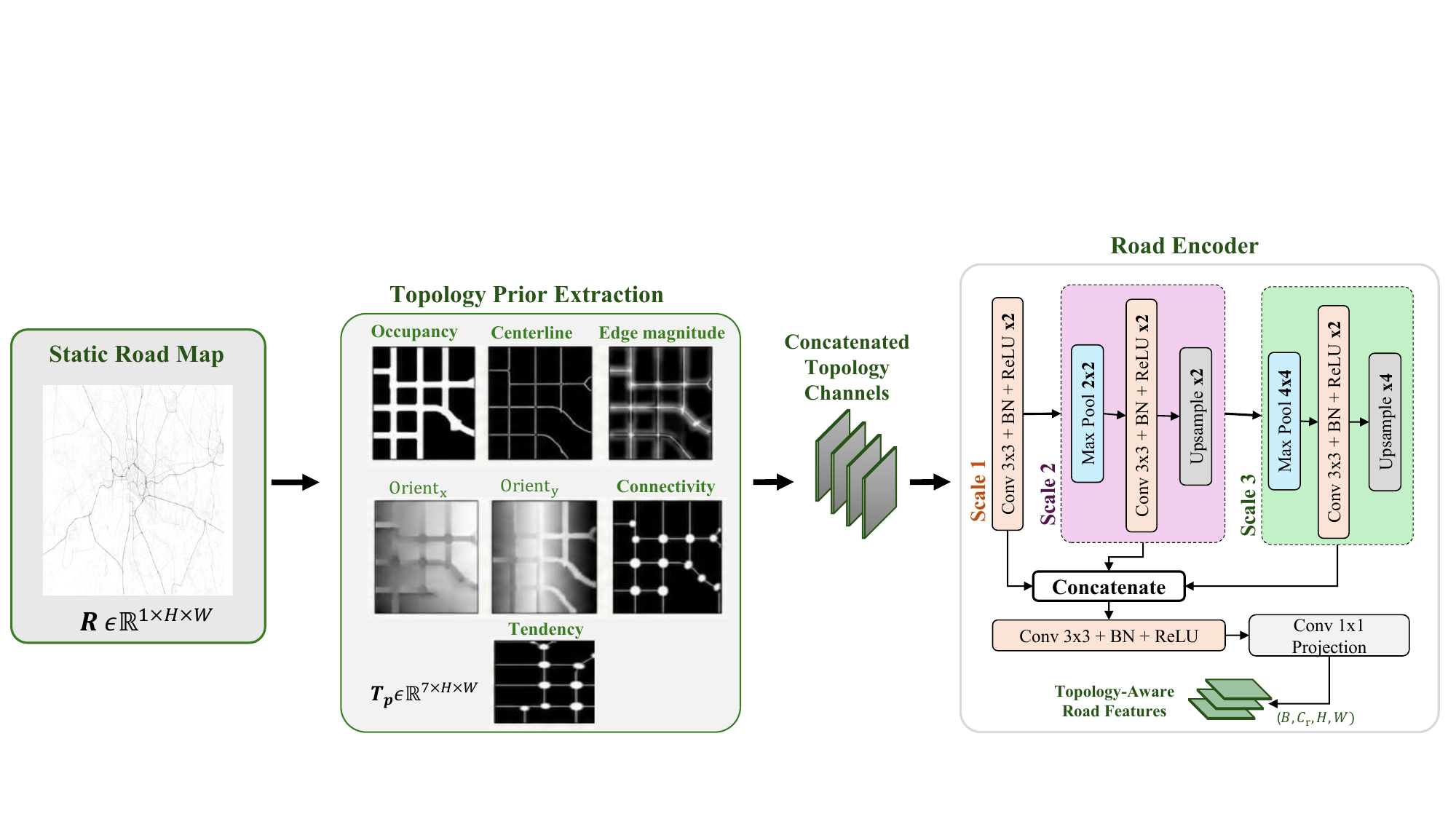}
    \caption{Detailed architecture of the topology-aware road representation module. The static road map is transformed into multiple structural prior channels, including occupancy, centerline, edge magnitude, orientation, connectivity, and intersection tendency, before being encoded through multi-scale spatial branches to produce topology-aware road features.}
    \label{fig:road}
\end{figure}

A binary road mask indicates whether each grid cell belongs to the road network, but it does not fully describe the structural properties that influence traffic movement. For traffic movie prediction, this limitation is important because future traffic states are shaped not only by road presence, but also by road orientation, local connectivity, centerline structure, and intersection geometry. Therefore, the proposed framework first transforms the static road map into a richer topology-aware representation (shown in Fig~\ref{fig:road}) before fusing it with dynamic traffic features.

Given the static road map be denoted as
\begin{equation}
\mathbf{R} \in \mathbb{R}^{1 \times H \times W},
\end{equation}
where \(H\) and \(W\) are the spatial height and width of the traffic grid. From \(\mathbf{R}\), a topology prior \(\mathbf{T}_p\) is constructed by extracting complementary structural cues:
\begin{equation}
\mathbf{T}_p =
\left[
\mathbf{R}_{occ};
\mathbf{R}_{cen};
\mathbf{R}_{edge};
\mathbf{R}_{ori}^{x};
\mathbf{R}_{ori}^{y};
\mathbf{R}_{con};
\mathbf{R}_{int}
\right],
\end{equation}
where \(\mathbf{R}_{occ}\) represents road occupancy, \(\mathbf{R}_{cen}\) is a centerline proximity proxy, \(\mathbf{R}_{edge}\) captures road boundary strength, \(\mathbf{R}_{ori}^{x}\) and \(\mathbf{R}_{ori}^{y}\) encode local orientation cues, \(\mathbf{R}_{con}\) represents local connectivity density, and \(\mathbf{R}_{int}\) approximates intersection tendency. Thus,
\begin{equation}
\mathbf{T}_p \in \mathbb{R}^{C_p \times H \times W},
\end{equation}
where \(C_p=7\) in this study.

The edge and orientation cues are computed from spatial gradients of the static road map. Specifically, Sobel filters are applied along the horizontal and vertical directions to obtain
\begin{equation}
\mathbf{G}_x = \mathcal{S}_x(\mathbf{R}), 
\qquad
\mathbf{G}_y = \mathcal{S}_y(\mathbf{R}),
\end{equation}
where \(\mathcal{S}_x(\cdot)\) and \(\mathcal{S}_y(\cdot)\) denote Sobel operations. The edge magnitude is then computed as
\begin{equation}
\mathbf{R}_{edge} =
\sqrt{\mathbf{G}_x^2 + \mathbf{G}_y^2 + \epsilon},
\end{equation}
while the orientation components are obtained as
\begin{equation}
\mathbf{R}_{ori}^{x} =
\frac{\mathbf{G}_x}{\sqrt{\mathbf{G}_x^2 + \mathbf{G}_y^2 + \epsilon}},
\qquad
\mathbf{R}_{ori}^{y} =
\frac{\mathbf{G}_y}{\sqrt{\mathbf{G}_x^2 + \mathbf{G}_y^2 + \epsilon}}.
\end{equation}
Here, \(\epsilon\) is a small constant used for numerical stability.

To capture broader road structure, local connectivity is estimated using spatial average pooling:
\begin{equation}
\mathbf{R}_{con} = \mathcal{A}_{k}(\mathbf{R}),
\end{equation}
where \(\mathcal{A}_{k}(\cdot)\) denotes average pooling with kernel size \(k\). Intersection tendency is estimated by combining local connectivity with curvature-like responses from a Laplacian filter:
\begin{equation}
\mathbf{R}_{int}
=
\mathbf{R}_{con}
\odot
\left|
\mathcal{L}(\mathbf{R})
\right|,
\end{equation}
where \(\mathcal{L}(\cdot)\) is the Laplacian operator and \(\odot\) denotes element-wise multiplication. This gives stronger responses near regions where road connectivity and local structural changes are both high.

After constructing the topology prior, a multi-scale convolutional road encoder is used to learn higher-level road features:
\begin{equation}
\mathbf{F}^{r} = \phi_r(\mathbf{T}_p),
\end{equation}
where \(\phi_r(\cdot)\) denotes the topology-aware road encoder and \(\mathbf{F}^{r}\) is the learned road feature map. The encoder processes \(\mathbf{T}_p\) at multiple spatial scales. The first branch operates at the original resolution to preserve fine road geometry, while the second and third branches operate on downsampled versions of the topology prior to capture broader connectivity, corridor continuity, and intersection-level structure. The lower-resolution features are then upsampled to the original grid size and concatenated with the full-resolution features:
\begin{equation}
\mathbf{F}^{r}
=
\psi_r
\left(
\left[
\phi_1(\mathbf{T}_p);
\mathcal{U}_2(\phi_2(\mathcal{D}_2(\mathbf{T}_p)));
\mathcal{U}_4(\phi_3(\mathcal{D}_4(\mathbf{T}_p)))
\right]
\right),
\end{equation}
where \(\phi_1(\cdot)\), \(\phi_2(\cdot)\), and \(\phi_3(\cdot)\) are convolutional encoding branches, \(\mathcal{D}_2(\cdot)\) and \(\mathcal{D}_4(\cdot)\) denote downsampling by factors of 2 and 4, \(\mathcal{U}_2(\cdot)\) and \(\mathcal{U}_4(\cdot)\) denote upsampling back to the original resolution, and \(\psi_r(\cdot)\) is a convolutional fusion function. The resulting feature map is
\begin{equation}
\mathbf{F}^{r} \in \mathbb{R}^{C_f \times H \times W},
\end{equation}
where \(C_f\) is the road feature dimension used by the downstream fusion module.

This representation allows the forecasting network to distinguish different road structures that may have different traffic behaviors. For example, a straight road segment, a turning region, and an intersection may all be valid road cells, but they influence traffic movement differently. By encoding occupancy, orientation, connectivity, and intersection-related information, the topology-aware road representation provides a structural prior that helps the downstream model generate forecasts that are not only numerically accurate, but also aligned with the physical road network.

\vspace{0.9em}
\subsubsection{Multi-Horizon Temporal Traffic Encoder}

\begin{figure*}
    \centering
    \includegraphics[width=0.8\linewidth]{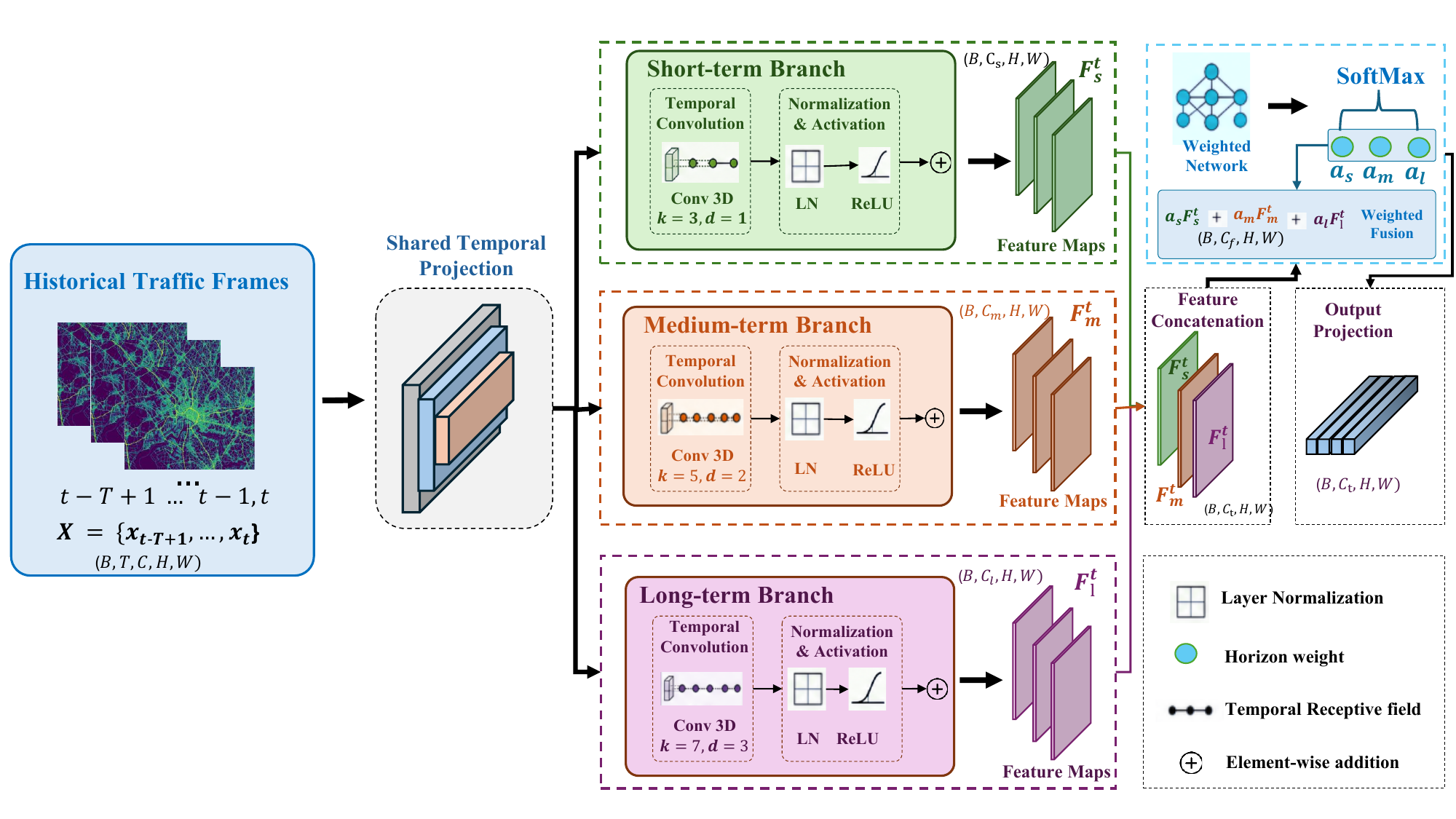}
    \caption{Detailed architecture of the multi-horizon temporal traffic encoder. Historical traffic sequences are processed through parallel temporal convolution branches with different kernel sizes and dilation rates to capture short-term, intermediate-term, and long-term traffic dynamics before feature fusion.}
    \label{fig:multi_encoder}
\end{figure*}

While the topology-aware road representation provides structural context, accurate traffic forecasting also requires a strong representation of how traffic evolves over time. Traffic dynamics do not occur at a single temporal scale. Short-term variations may reflect sudden speed changes, queue dissipation, or local fluctuations, while longer-term patterns may reflect congestion buildup, propagation, and broader directional flow transitions. To capture these different temporal behaviors, the proposed framework introduces a multi-horizon temporal traffic encoder as illustrated in Fig~\ref{fig:multi_encoder}.

Let the historical traffic sequence be denoted as
$\mathbf{X} \in \mathbb{R}^{B \times C \times T_{in} \times H \times W}$
where \(B\) is the batch size, \(C\) is the number of traffic channels, \(T_{in}\) is the number of observed frames, and \(H \times W\) is the spatial grid size. The encoder first projects the input sequence into a shared spatiotemporal feature representation,
\begin{equation}
\mathbf{F}_{0} = \phi_{0}(\mathbf{X}),
\end{equation}
where \(\phi_{0}(\cdot)\) denotes the initial 3D convolutional projection.

To model traffic evolution at different temporal horizons, \(\mathbf{F}_{0}\) is passed through three parallel temporal convolution branches with different kernel sizes and dilation rates. The short-horizon branch uses a smaller temporal receptive field to capture immediate local changes, the intermediate-horizon branch uses a moderate receptive field to model evolving traffic propagation, and the long-horizon branch uses a larger dilated receptive field to capture broader temporal trends. These branches are defined as
\begin{equation}
\mathbf{F}^{s}_{t} = \phi_{s}^{k_s,d_s}(\mathbf{F}_{0}), \qquad
\mathbf{F}^{m}_{t} = \phi_{m}^{k_m,d_m}(\mathbf{F}_{0}), \qquad
\mathbf{F}^{l}_{t} = \phi_{l}^{k_l,d_l}(\mathbf{F}_{0}),
\end{equation}
where \(\mathbf{F}^{s}_{t}\), \(\mathbf{F}^{m}_{t}\), and \(\mathbf{F}^{l}_{t}\) denote the short-, intermediate-, and long-horizon temporal features, respectively. The terms \(k_s\), \(k_m\), and \(k_l\) represent the temporal kernel sizes, while \(d_s\), \(d_m\), and \(d_l\) represent the corresponding dilation rates.

The effective temporal receptive field of each branch is given by
\begin{equation}
R_i = 1 + (k_i - 1)d_i, \qquad i \in \{s,m,l\},
\end{equation}
where \(R_i\) indicates how many temporal positions are covered by the corresponding branch. In this design, the short branch uses the smallest receptive field, the intermediate branch expands the temporal context, and the long branch captures wider dependencies across the observation window.

The multi-horizon features are then integrated through channel-wise concatenation followed by a learnable fusion projection:
\begin{equation}
\mathbf{F}^{temp}_{t}
=
\psi_t
\left(
\left[
\mathbf{F}^{s}_{t};
\mathbf{F}^{m}_{t};
\mathbf{F}^{l}_{t}
\right]
\right),
\end{equation}
where \([\cdot;\cdot;\cdot]\) denotes concatenation along the channel dimension, \(\psi_t(\cdot)\) denotes the fusion projection, and
\begin{equation}
\mathbf{F}^{temp}_{t} \in \mathbb{R}^{B \times C_t \times H \times W}
\end{equation}
is the final temporal traffic representation.

This design allows the encoder to capture immediate fluctuations, intermediate congestion propagation, and longer-range temporal trends within one unified representation. The resulting feature \(\mathbf{F}^{temp}_{t}\) provides the dynamic traffic context that is later aligned with the topology-aware road feature in the direction-aware road-traffic fusion module.

\vspace{0.9em}
\subsubsection{Direction-Aware Road-Traffic Fusion}

\begin{figure*}
    \centering
    \includegraphics[width=0.8\linewidth]{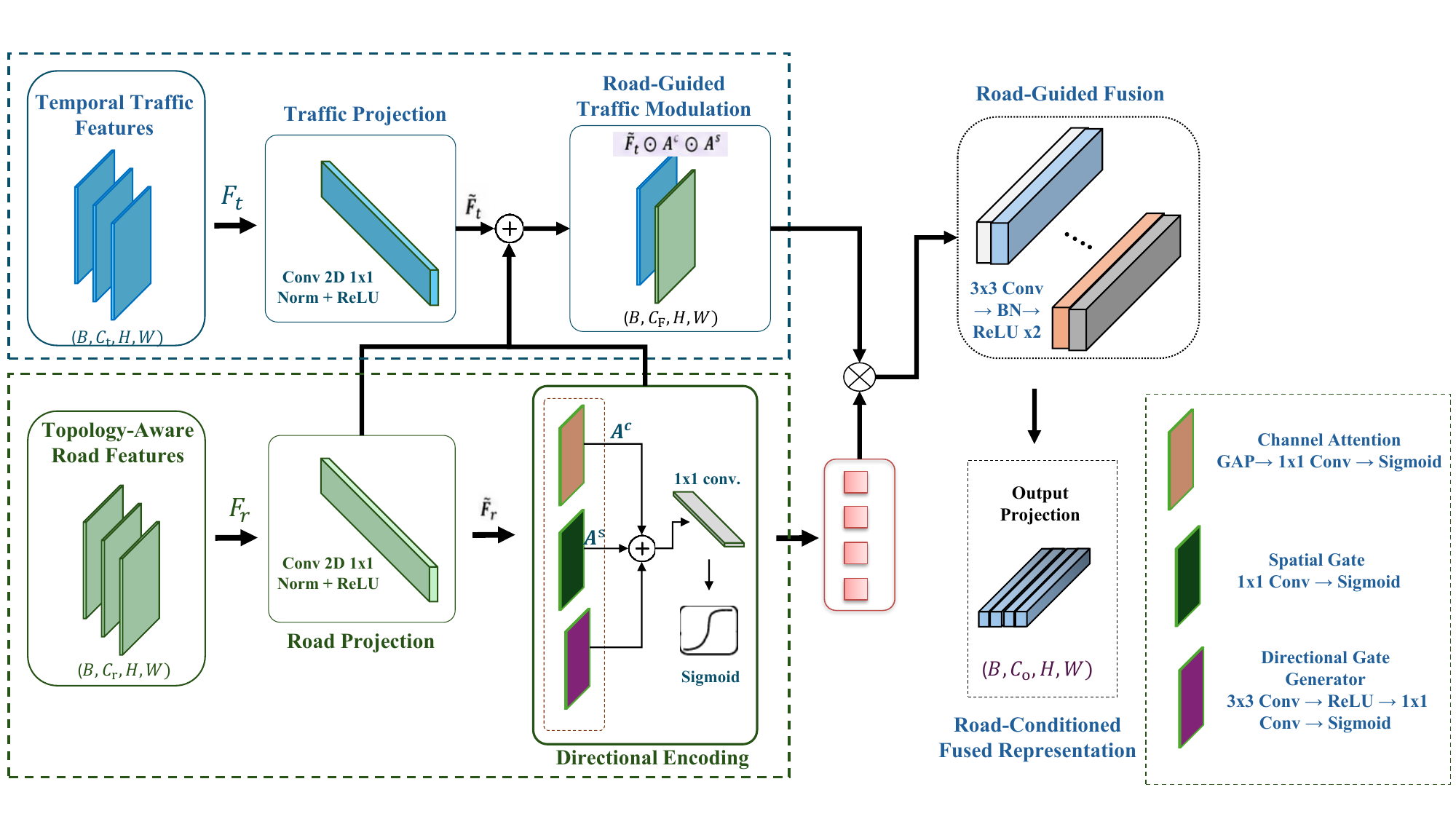}
    \caption{Detailed architecture of the direction-aware road-traffic fusion module. The topology-aware road feature guides channel, spatial, and direction-aware modulation of the traffic representation, enabling the fused feature to reflect both dynamic traffic evolution and local road structure.}
    \label{fig:directional}
\end{figure*}

After obtaining the topology-aware road feature $\mathbf{F}^{r}$ and the multi-horizon traffic feature $\mathbf{F}^{temp}_{t}$, the proposed framework which is illustrated in Fig.~\ref{fig:directional} integrates them through a direction-aware road-traffic fusion module. The purpose of this module is to use road structure to guide how dynamic traffic features are emphasized across spatial locations, feature channels, and movement directions. This is important because traffic on a straight road segment, an intersection, and a turning region should not be interpreted in the same way, even if their traffic intensities appear similar.

Given the traffic and road features denoted as

$\mathbf{F}^{temp}_{t} \in \mathbb{R}^{B \times C_t \times H \times W}$,
\qquad
$\mathbf{F}^{r} \in \mathbb{R}^{B \times C_r \times H \times W}$,

where $B$ is the batch size, $C_t$ and $C_r$ are the traffic and road feature dimensions, and $H \times W$ is the spatial grid size. The road feature is first projected into the same feature space as the traffic representation:
\begin{equation}
\widetilde{\mathbf{F}}^{r}
=
P_r(\mathbf{F}^{r}),
\end{equation}
where $P_r(\cdot)$ denotes a $1 \times 1$ convolution followed by normalization and activation.

The projected road feature is then used to generate a channel attention vector that controls which traffic feature channels should be emphasized:
\begin{equation}
\mathbf{A}^{c}
=
\sigma
\left(
P_c
\left(
\mathrm{GAP}(\widetilde{\mathbf{F}}^{r})
\right)
\right),
\end{equation}
where $\mathrm{GAP}(\cdot)$ denotes global average pooling, $P_c(\cdot)$ is a channel attention projection, and $\sigma(\cdot)$ is the sigmoid activation. In parallel, a spatial gate is generated from the road feature:
\begin{equation}
\mathbf{A}^{s}
=
\sigma
\left(
P_s(\mathbf{F}^{r})
\right),
\end{equation}
where $P_s(\cdot)$ denotes a $1 \times 1$ convolution. The traffic feature is then modulated by both channel and spatial road guidance:
\begin{equation}
\widetilde{\mathbf{F}}^{temp}_{t}
=
\mathbf{F}^{temp}_{t}
\odot
\mathbf{A}^{c}
\odot
\mathbf{A}^{s},
\end{equation}
where $\odot$ denotes element-wise multiplication.

To further encode road-directional constraints, the module predicts a direction-aware gate from the road feature:
\begin{equation}
\mathbf{G}^{dir}
=
\sigma
\left(
P_{dir}(\mathbf{F}^{r})
\right),
\end{equation}
where $P_{dir}(\cdot)$ is implemented using a $3 \times 3$ convolution, activation, and a $1 \times 1$ convolution. The resulting direction gate is
\begin{equation}
\mathbf{G}^{dir} \in \mathbb{R}^{B \times 4 \times H \times W},
\end{equation}
where the four channels represent learned direction-sensitive road responses. These responses are not manually assigned to fixed directions; instead, they are learned from the road topology and help the model distinguish local directional structures such as straight segments, turns, and intersections.

The road-guided traffic feature, projected road feature, and direction gate are then concatenated:
\begin{equation}
\mathbf{F}^{cat}
=
\left[
\widetilde{\mathbf{F}}^{temp}_{t};
\widetilde{\mathbf{F}}^{r};
\mathbf{G}^{dir}
\right].
\end{equation}
This combined representation is passed through convolutional fusion layers to obtain a refined road-traffic interaction feature:
\begin{equation}
\mathbf{F}^{ref}
=
P_f(\mathbf{F}^{cat}),
\end{equation}
where $P_f(\cdot)$ denotes the fusion convolutional function.

Finally, a residual connection is added from the original traffic representation to preserve dynamic traffic information while incorporating road-conditioned refinement:
\begin{equation}
\mathbf{F}^{rt}_{t}
=
\mathbf{F}^{temp}_{t}
+
\mathbf{F}^{ref}.
\end{equation}
The resulting feature

$\mathbf{F}^{rt}_{t} \in \mathbb{R}^{B \times C_f \times H \times W}$
is the road-conditioned traffic representation passed to the progressive forecasting decoder.

Through this design, the module does not simply concatenate road and traffic features. Instead, it uses the road representation to modulate traffic channels, control spatial emphasis, introduce learned direction-sensitive gates, and refine the traffic representation through residual fusion. This allows the downstream decoder to generate future traffic maps that are more consistent with the physical and directional structure of the road network.

\vspace{0.9em}
\subsubsection{Progressive Multi-Step Forecast Decoder}

After obtaining the road-conditioned traffic representation $\mathbf{F}^{rt}_{t}$, the model generates the future traffic sequence using a progressive multi-step forecast decoder. This design is motivated by the fact that future traffic states are temporally dependent, meaning that the prediction process should preserve continuity across the forecasting horizon instead of producing all future frames as unrelated outputs.

Let the fused representation be
$\mathbf{F}^{rt}_{t} \in \mathbb{R}^{B \times C_f \times H \times W}$,
where $B$ is the batch size, $C_f$ is the fused feature dimension, and $H \times W$ is the spatial grid size. The decoder first refines this representation using convolutional context projection:
\begin{equation}
\mathbf{C}_{t} = \phi_c(\mathbf{F}^{rt}_{t}),
\end{equation}
where $\phi_c(\cdot)$ denotes the context projection function and $\mathbf{C}_{t}$ is the shared spatial context used for all forecast steps.

To initialize the progressive decoding process, global average pooling is applied to $\mathbf{C}_{t}$, followed by a linear projection:
\begin{equation}
\mathbf{z}_{0} = \mathrm{GAP}(\mathbf{C}_{t}), \qquad
\mathbf{h}_{0} = \tanh(\mathbf{W}_{h}\mathbf{z}_{0}),
\end{equation}
where $\mathbf{z}_{0}$ is the global context vector and $\mathbf{h}_{0}$ is the initial hidden forecasting state.

For each future step $k=1,\ldots,T_{out}$, the hidden state is updated using a gated recurrent unit:
\begin{equation}
\mathbf{h}_{k} = \mathrm{GRUCell}(\mathbf{z}_{k-1}, \mathbf{h}_{k-1}).
\end{equation}
The updated hidden state is then projected back into a spatial feature embedding and added to the shared context:
\begin{equation}
\mathbf{E}_{k} = \mathrm{reshape}(\mathbf{W}_{e}\mathbf{h}_{k}),
\qquad
\mathbf{S}_{k} = \mathbf{C}_{t} + \mathbf{E}_{k},
\end{equation}
where $\mathbf{E}_{k} \in \mathbb{R}^{B \times C_f \times 1 \times 1}$ is broadcast over the spatial grid, and $\mathbf{S}_{k}$ is the step-specific decoding feature.

Each step-specific feature is then passed through shared convolutional layers and two separate prediction heads:
\begin{equation}
\mathbf{Q}_{k} = \phi_s(\mathbf{S}_{k}),
\end{equation}
\begin{equation}
\hat{\mathbf{Y}}^{vol}_{k} = \phi_{vol}(\mathbf{Q}_{k}), 
\qquad
\hat{\mathbf{Y}}^{spd}_{k} = \phi_{spd}(\mathbf{Q}_{k}),
\end{equation}
where $\hat{\mathbf{Y}}^{vol}_{k} \in \mathbb{R}^{B \times 4 \times H \times W}$ and $\hat{\mathbf{Y}}^{spd}_{k} \in \mathbb{R}^{B \times 4 \times H \times W}$ denote the predicted directional volume and speed channels, respectively. These two outputs are interleaved to form the eight-channel traffic prediction:
\begin{equation}
\hat{\mathbf{Y}}_{k}
=
\mathrm{Interleave}
\left(
\hat{\mathbf{Y}}^{vol}_{k},
\hat{\mathbf{Y}}^{spd}_{k}
\right),
\end{equation}
with $\hat{\mathbf{Y}}_{k} \in \mathbb{R}^{B \times 8 \times H \times W}.$

After each step, the decoder updates the next recurrent input using the pooled step-specific feature:
\begin{equation}
\mathbf{z}_{k} = \mathrm{GAP}(\mathbf{S}_{k}).
\end{equation}
Finally, all predicted steps are stacked to obtain the future traffic sequence:
\begin{equation}
\hat{\mathbf{Y}}
=
\left[
\hat{\mathbf{Y}}_{1},
\hat{\mathbf{Y}}_{2},
\ldots,
\hat{\mathbf{Y}}_{T_{out}}
\right]
\in
\mathbb{R}^{B \times T_{out} \times 8 \times H \times W}.
\end{equation}

This progressive design allows the decoder to maintain a recurrent forecasting state across future steps while preserving the shared spatial context learned from the road-conditioned representation. By predicting directional volume and speed through separate heads, the decoder also preserves the semantic organization of the traffic channels before forming the final multi-channel future traffic map.

\vspace{0.9em}
\subsection{Structure-Consistent Learning Objective}

The proposed framework is trained using a structure-consistent objective that encourages accurate, road-aligned, temporally coherent, and spatially stable forecasts. A standard regression loss is useful for reducing numerical error, but it treats all grid cells uniformly and does not explicitly emphasize valid road regions where meaningful traffic activity occurs. Therefore, the training objective combines the main prediction loss with road-weighted structure consistency, temporal consistency, and spatial edge consistency.

Let $\mathbf{Y}$ and $\hat{\mathbf{Y}}$ denote the ground-truth and predicted future traffic sequences, respectively, where both are defined over $T_{out}$ future steps. The overall objective is formulated as
\begin{equation}
\mathcal{L}
=
\mathcal{L}_{pred}
+
\lambda_s \mathcal{L}_{struct}
+
\lambda_t \mathcal{L}_{temp}
+
\lambda_e \mathcal{L}_{edge},
\end{equation}
where $\mathcal{L}_{pred}$ is the main prediction loss, $\mathcal{L}_{struct}$ is the road-weighted structure-consistency loss, $\mathcal{L}_{temp}$ is the temporal-consistency loss, $\mathcal{L}_{edge}$ is the spatial edge-consistency loss, and $\lambda_s$, $\lambda_t$, and $\lambda_e$ are balancing coefficients.

The prediction loss measures the average squared difference between the predicted and ground-truth traffic maps:
\begin{equation}
\mathcal{L}_{pred}
=
\frac{1}{N}
\sum_{i=1}^{N}
\left(
\hat{\mathbf{Y}}_i - \mathbf{Y}_i
\right)^2,
\end{equation}
where $N$ denotes the total number of evaluated elements across future steps, channels, and spatial locations.

To make the model pay stronger attention to valid road regions, the structure-consistency loss applies a road-dependent weight to the squared prediction error:
\begin{equation}
\mathcal{L}_{struct}
=
\frac{1}{N}
\sum_{i=1}^{N}
w_i
\left(
\hat{\mathbf{Y}}_i - \mathbf{Y}_i
\right)^2,
\end{equation}
where
\begin{equation}
w_i =
\begin{cases}
\gamma, & \text{if } \mathbf{R}_i > \tau,\\
1, & \text{otherwise}.
\end{cases}
\end{equation}
Here, $\mathbf{R}$ is the static road map, $\tau$ is the road-mask threshold, and $\gamma$ is the road-region weight. This term gives higher importance to prediction errors on valid road cells while still retaining background supervision.

To encourage smooth traffic evolution across the forecast horizon, the temporal-consistency loss compares the frame-to-frame changes of the prediction with those of the ground truth:
\begin{equation}
\mathcal{L}_{temp}
=
\frac{1}{T_{out}-1}
\sum_{t=1}^{T_{out}-1}
\left\|
(\hat{\mathbf{Y}}_{t+1}-\hat{\mathbf{Y}}_{t})
-
(\mathbf{Y}_{t+1}-\mathbf{Y}_{t})
\right\|_2^2.
\end{equation}

Finally, the edge-consistency loss encourages the spatial gradients of the predicted maps to follow the spatial gradients of the ground truth:
\begin{equation}
\mathcal{L}_{edge}
=
\left\|
\nabla_x \hat{\mathbf{Y}} - \nabla_x \mathbf{Y}
\right\|_1
+
\left\|
\nabla_y \hat{\mathbf{Y}} - \nabla_y \mathbf{Y}
\right\|_1,
\end{equation}
where $\nabla_x$ and $\nabla_y$ denote horizontal and vertical spatial gradients. This term helps reduce overly smooth or spatially diffuse forecasts by preserving local traffic-map structure. Together, these loss terms guide the model to minimize numerical prediction error while maintaining road alignment, temporal continuity, and spatial structure across the predicted traffic sequence.

% Together, these three terms guide the model to generate forecasts that are accurate, structurally meaningful, and temporally coherent. As a result, the learning objective is aligned with the main goal of the proposed framework, which is to predict future traffic states while respecting both the dynamic evolution of traffic and the physical constraints imposed by the road network.

\section{Experiments}
\label{experiment}
This section describes the experimental setup used to evaluate the proposed framework. It includes the dataset, preprocessing procedure, implementation details, evaluation metrics, and comparative baseline methods.

\subsection{Dataset Description}

The experiments are conducted on the Traffic4cast dataset \cite{pmlr-v176-eichenberger22a}, a city-wide traffic forecasting benchmark built from aggregated GPS probe data. The dataset contains dynamic traffic maps and static road maps for each city. In this study, Berlin, Antwerp, and Moscow are used for the main same-city experiments, while Chicago and Bangkok are used as unseen target cities for cross-city generalization.

Each dynamic traffic file contains grid-based traffic maps with eight channels, representing directional traffic volume and speed. The static road map is stored separately for each city and is normalized to the range \([0,1]\) before being used as the structural input to the model. Dynamic traffic tensors are arranged as \((T, H, W, 8)\) in the raw files and converted to \((T, C, H, W)\) during training, where \(C=8\). The static map is reduced to one channel when needed and returned as \((1,H,W)\). This follows the data format used in the implementation. For the same-city experiments, the files of each city are split into 70\% training, 15\% validation, and 15\% testing. For cross-city generalization, the model is trained using only the source cities, namely Berlin, Antwerp, and Moscow, and then evaluated directly on Chicago and Bangkok. 

\subsection{Implementation Details}

The proposed framework is implemented in PyTorch and trained on a single NVIDIA A40 GPU. All experiments use 12 historical traffic frames to predict the next 12 future frames, with samples generated using a sliding window stride of 6. Dynamic traffic maps are normalized channel-wise using statistics computed from the training set, while the static road maps are normalized to \([0,1]\) and reduced to one channel before topology-prior extraction. AdamW is used for optimization with an initial learning rate of \(1 \times 10^{-3}\), weight decay of \(1 \times 10^{-4}\), and a cosine annealing learning-rate schedule. The batch size is set to 8, and the model is trained for 50 epochs. The base channel size is set to 32, and the hidden dimension of the progressive decoder is set to 128. Gradient clipping is applied with a maximum norm of 1.0, and the random seed is fixed at 42 for reproducibility. The training objective combines the main prediction loss with structure-consistency terms for road alignment, temporal consistency, and spatial edge consistency. The road-region weight is set to 5.0, while the structural, temporal, and edge consistency coefficients are set to 0.5, 0.2, and 0.1, respectively. The checkpoint with the lowest validation loss is selected for final testing.

\subsection{Evaluation Metrics}

Model performance is evaluated using mean absolute error (MAE), mean squared error (MSE), and root mean square error (RMSE) between the predicted and ground-truth traffic maps. These metrics are computed over all predicted frames, channels, and spatial locations to measure the accuracy of the multi-step traffic forecasts.

Specifically, let \(\hat{Y}\) and \(Y\) denote the predicted and ground-truth traffic sequences, respectively, and let \(N\) be the total number of evaluated elements over all forecast steps, channels, and spatial positions. The MAE is defined as
\begin{equation}
\mathrm{MAE} = \frac{1}{N} \sum_{i=1}^{N} \left| \hat{Y}_i - Y_i \right|,
\end{equation}
the MSE is given by
\begin{equation}
\mathrm{MSE} = \frac{1}{N} \sum_{i=1}^{N} \left( \hat{Y}_i - Y_i \right)^2,
\end{equation}
and the RMSE is computed as
\begin{equation}
\mathrm{RMSE} = \sqrt{ \frac{1}{N} \sum_{i=1}^{N} \left( \hat{Y}_i - Y_i \right)^2 }.
\end{equation}

Lower values of these metrics indicate better agreement between the predicted traffic states and the corresponding ground-truth observations.

\subsection{Baseline Methods}

To evaluate the effectiveness of the proposed framework, comparisons are made against several representative traffic forecasting methods.

\vspace{0.6em}
\begin{itemize}
   \item \textit{Historical Average}: Historical Average predicts future traffic by computing the mean of the input sequence, assuming traffic conditions remain constant over the prediction horizon.

    \vspace{0.6em}
    \item \textit{Multi-task Transfer Learning \cite{lu2021learning}}: This method learns a shared forecasting model across multiple cities using a multi-task learning strategy to improve generalization across related traffic domains.

\vspace{0.6em}
    \item \textit{UNet \cite{choi2020utilizing}}: This baseline uses a U-Net architecture to predict future traffic maps from historical observations through an encoder-decoder structure with skip connections.

\vspace{0.6em}

    \item \textit{U-Net with Temporal Domain Adaptation \cite{konyakhin2021solving}}: This approach combines a U-Net forecasting backbone with temporal domain adaptation to reduce the effect of time-dependent distribution shifts.

\vspace{0.6em}

    \item \textit{Traffic Forecasting on Traffic Movie Snippets \cite{wiedemann2021traffic}}: This method treats traffic forecasting as a traffic-movie prediction task and learns spatiotemporal patterns directly from short sequences of traffic frames.

\vspace{0.6em}

    \item \textit{3DResNet and Sparse-UNet \cite{wang2021traffic4cast}}: This baseline combines 3D residual convolutions for temporal feature learning with a sparse U-Net for spatial prediction over large traffic grids.

\vspace{0.6em}

    \item \textit{A Graph-based U-Net \cite{hermes2022graph}}: This method integrates graph-based road-network modeling with a U-Net architecture to capture both relational and multi-scale spatial dependencies.

\vspace{0.6em}

    \item \textit{SwinUnet3D \cite{bojesomo2022swinunet3d}}: This baseline uses a hierarchical transformer architecture with shifted-window attention to model spatiotemporal traffic dynamics.
\vspace{0.6em}

    \item \textit{Dual Encoding U-Net \cite{santokhi2021dual}}: This approach uses dual encoding branches to capture complementary traffic representations before combining them within a U-Net-style prediction framework.
\end{itemize}

\section{Results}
\label{results}
This section presents the performance of the proposed framework under the experimental setting described in Section \ref{experiment}. Quantitative results are first reported to compare the proposed method with representative baseline approaches, followed by further analysis of the contribution of the proposed components and the quality of the predicted traffic maps.

\subsection{Quantitative Results}

The city-specific quantitative results are reported in Table~\ref{tab:city_specific}. For each city, the predicted traffic sequences on the held-out test set are compared with their corresponding ground-truth future sequences. MAE, MSE, and RMSE are then computed by aggregating the prediction errors over all forecast steps, channels, and spatial locations within that city. In this way, each metric summarizes forecasting error across the full prediction horizon, with lower values indicating better performance.

The results in Table~\ref{tab:city_specific} show a clear and consistent pattern across Berlin, Antwerp, and Moscow. The Historical Average baseline records the highest errors in all three cities because it does not learn traffic dynamics and only reflects average past behavior. Once learning-based methods are introduced, the errors decrease substantially, showing that reliable traffic forecasting depends on learning both spatial structure and temporal evolution from data. Among the learned baselines, Multi-task Transfer Learning \cite{lu2021learning} gives the strongest and most consistent performance across the three cities. This is expected because shared training across multiple cities helps the model learn more transferable traffic patterns. Even so, the proposed framework still achieves lower errors in every city. In Berlin, the MAE, MSE, and RMSE are reduced by 10.5\%, 9.4\%, and 4.8\%, respectively. In Antwerp, the reductions increase to 14.9\%, 10.8\%, and 5.5\%. In Moscow, the corresponding gains are 9.5\%, 9.7\%, and 5.0\%. UNet \cite{choi2020utilizing} follows closely in all three cities, showing that encoder-decoder reconstruction with skip connections is effective for recovering spatial traffic maps. However, its representation remains mainly reconstruction-driven and does not explicitly account for road structure or directional traffic behavior. Relative to UNet, the proposed framework reduces MAE, MSE, and RMSE by 14.7\%, 12.8\%, and 6.6\% in Berlin, by 19.1\%, 14.2\%, and 7.4\% in Antwerp, and by 14.2\%, 12.8\%, and 6.6\% in Moscow. U-Net with Temporal Domain Adaptation \cite{konyakhin2021solving} also remains competitive because temporal adaptation helps reduce distribution shift across time. Still, this alone does not fully capture the structural and directional complexity of traffic evolution. Compared with this baseline, the proposed model reduces MAE, MSE, and RMSE by 22.1\%, 21.9\%, and 11.6\% in Berlin, by 25.8\%, 23.1\%, and 12.3\% in Antwerp, and by 21.5\%, 21.2\%, and 11.2\% in Moscow. 

The remaining baselines, namely Traffic Movie Snippets \cite{wiedemann2021traffic}, 3DResNet with Sparse-UNet \cite{wang2021traffic4cast}, Graph-based U-Net \cite{hermes2022graph}, SwinUnet3D \cite{bojesomo2022swinunet3d}, and Dual Encoding U-Net \cite{santokhi2021dual}, all strengthen forecasting through richer temporal modeling, graph reasoning, transformer attention, or dual-branch encoding. However, their errors remain consistently higher than those of the proposed framework across all three cities. Relative to these methods, the proposed model reduces MAE by 14.7\% to 30.0\% in Berlin, 19.0\% to 34.1\% in Antwerp, and 14.2\% to 29.9\% in Moscow while the corresponding MSE reductions range from 11.3\% to 29.2\% in Berlin, 14.2\% to 30.6\% in Antwerp, and 12.8\% to 29.5\% in Moscow respectively. These gains show that the proposed framework models traffic more effectively than the baselines. While the compared methods mainly emphasize transfer learning, spatial reconstruction, temporal adaptation, graph reasoning, or attention separately, the proposed model combines road structure and traffic evolution more directly. This helps it understand where traffic can occur, how it changes over time, and how future states should be generated, leading to lower forecasting error across all evaluated cities.

\begin{table*}[!t]
\centering
\caption{City-specific performance comparison on same-city traffic forecasting. Models are trained and tested on the same city. Best results per city are in \textcolor{red}{\textbf{Red}}, second best are \underline{underlined}.}
\label{tab:city_specific}
\renewcommand{\arraystretch}{1.15}
\resizebox{\textwidth}{!}{%
\begin{tabular}{l | r r r | r r r | r r r}
\toprule
\multirow{2}{*}{\textbf{Method}} & \multicolumn{3}{c|}{\textbf{BERLIN}} & \multicolumn{3}{c|}{\textbf{ANTWERP}} & \multicolumn{3}{c}{\textbf{MOSCOW}} \\
 & MAE$\downarrow$ & MSE$\downarrow$ & RMSE$\downarrow$ & MAE$\downarrow$ & MSE$\downarrow$ & RMSE$\downarrow$ & MAE$\downarrow$ & MSE$\downarrow$ & RMSE$\downarrow$ \\
\midrule
Historical Average & 0.2324 & 1.1583 & 1.0762 & 0.2156 & 1.0894 & 1.0438 & 0.2489 & 1.2167 & 1.1030 \\
Multi-task Transfer Learning \cite{lu2021learning} & \underline{0.0745} & \underline{0.4512} & \underline{0.6717} & \underline{0.0698} & \underline{0.4234} & \underline{0.6507} & \underline{0.0801} & \underline{0.4789} & \underline{0.6920} \\
UNet \cite{choi2020utilizing} & 0.0782 & 0.4689 & 0.6848 & 0.0734 & 0.4401 & 0.6634 & 0.0845 & 0.4956 & 0.7040 \\
U-Net + TDA \cite{konyakhin2021solving} & 0.0856 & 0.5234 & 0.7235 & 0.0801 & 0.4912 & 0.7008 & 0.0923 & 0.5489 & 0.7408 \\
Traffic Movie Snippets \cite{wiedemann2021traffic} & 0.0912 & 0.5567 & 0.7461 & 0.0856 & 0.5234 & 0.7235 & 0.0989 & 0.5901 & 0.7682 \\
3DResNet + Sparse-UNet \cite{wang2021traffic4cast} & 0.0889 & 0.5401 & 0.7349 & 0.0834 & 0.5067 & 0.7118 & 0.0967 & 0.5734 & 0.7572 \\
Graph-based U-Net \cite{hermes2022graph} & 0.0934 & 0.5678 & 0.7535 & 0.0878 & 0.5334 & 0.7304 & 0.1012 & 0.6012 & 0.7754 \\
SwinUnet3D \cite{bojesomo2022swinunet3d} & 0.0867 & 0.5289 & 0.7273 & 0.0812 & 0.4967 & 0.7048 & 0.0945 & 0.5623 & 0.7499 \\
Dual Encoding U-Net \cite{santokhi2021dual} & 0.0956 & 0.5789 & 0.7609 & 0.0901 & 0.5445 & 0.7379 & 0.1034 & 0.6134 & 0.7832 \\
\midrule
\textbf{RCSNet (Ours)} & \textcolor{red}{\textbf{0.0667}} & \textcolor{red}{\textbf{0.4089}} & \textcolor{red}{\textbf{0.6395}} & \textcolor{red}{\textbf{0.0594}} & \textcolor{red}{\textbf{0.3776}} & \textcolor{red}{\textbf{0.6145}} & \textcolor{red}{\textbf{0.0725}} & \textcolor{red}{\textbf{0.4323}} & \textcolor{red}{\textbf{0.6575}} \\
\bottomrule
\end{tabular}%
}
\end{table*}

\subsection{Ablation Study}

To examine the contribution of each proposed component, the ablation study progressively builds the final framework from a simple baseline, as shown in Table~\ref{tab:ablation}. The baseline contains none of the proposed modules. TA-RR only adds the topology-aware road representation branch with a simple prediction head, while MHTE only adds the multi-horizon temporal encoder with the same simple prediction head. The TA-RR + MHTE variant combines both encoders using simple fusion. DAF is then introduced to replace this simple combination with direction-aware fusion, PMD adds progressive multi-step decoding, and SCL completes the framework by introducing the structure-consistent learning objective.

As shown in Table~\ref{tab:ablation}, a clear progression is observed across Berlin, Antwerp, and Moscow. Starting from the baseline, adding TA-RR reduces MAE by 12.7\%, 13.0\%, and 13.1\% in Berlin, Antwerp, and Moscow, respectively. The corresponding MSE reductions are 16.4\%, 16.9\%, and 16.3\%. This shows that explicit road structure already provides strong spatial guidance for forecasting. Adding MHTE alone also improves the baseline, but the gain is smaller, with MAE reductions of 7.3\%, 7.4\%, and 8.2\% across the three cities. This indicates that richer temporal modeling is useful, although the road branch provides the stronger first improvement. When TA-RR and MHTE are combined, the error drops further in every city. Relative to the baseline, this variant reduces MAE by 23.7\% in Berlin, 24.2\% in Antwerp, and 24.6\% in Moscow. The corresponding MSE reductions are 26.8\%, 27.3\%, and 26.7\%, while RMSE decreases by 14.4\%, 14.7\%, and 14.4\%. This confirms that structural road information and temporal traffic dynamics are complementary, and that better forecasting requires both.

A further improvement appears when DAF is introduced. Compared with the TA-RR + MHTE variant, DAF lowers MAE by 12.7\% in Berlin, 12.4\% in Antwerp, and 11.9\% in Moscow. MSE decreases by 13.2\%, 13.1\%, and 13.1\%, while RMSE drops by 5.5\%, 6.7\%, and 6.8\%, respectively. This shows that simply combining the two encoders is not enough, and that explicitly aligning road structure with traffic dynamics leads to a stronger representation. The addition of PMD reduces the error again across all three cities. Relative to the model with DAF, MAE decreases by 9.0\% in Berlin, 9.8\% in Antwerp, and 8.7\% in Moscow. The corresponding MSE reductions are 10.7\%, 11.1\%, and 10.4\%, while RMSE decreases by 5.5\%, 5.7\%, and 5.5\%. This indicates that future traffic states are predicted more effectively when they are generated progressively rather than through a simple direct head.

The final improvement comes from SCL. Compared with the PMD variant, adding the structure-consistent learning objective reduces MAE by 11.8\% in Berlin, 16.6\% in Antwerp, and 11.9\% in Moscow. The corresponding MSE reductions are 15.4\%, 18.3\%, and 15.1\%, while RMSE decreases by 8.4\%, 9.9\%, and 8.2\%, respectively. With this final addition, the full RCSNet achieves the best results in all three cities. Relative to the baseline, the full model reduces MAE by 46.5\% in Berlin, 50.0\% in Antwerp, and 46.5\% in Moscow while the corresponding MSE reductions are 52.0\%, 54.1\%, and 51.5\%.

These results show that no single component is responsible for the final performance. Instead, the gains accumulate as road-aware structural reasoning, stronger temporal encoding, more effective fusion, progressive decoding, and structure-consistent supervision are introduced step by step. This is why the full model gives the lowest error across all three cities.

\newcommand{\cmark}{\checkmark}
\newcommand{\xmark}{$\times$}

\begin{table*}[!t]
\centering
\caption{Ablation study of RCSNet components on same-city traffic forecasting. \cmark\ indicates the module is included; \xmark\ indicates it is excluded. Best results per city are in \textbf{bold}.}
\label{tab:ablation}
\renewcommand{\arraystretch}{1.15}
\resizebox{\textwidth}{!}{%
\begin{tabular}{l | c c c c c | r r r | r r r | r r r}
\toprule
\multirow{2}{*}{\textbf{Model Variant}} & \multicolumn{5}{c|}{\textbf{Module Components}} & \multicolumn{3}{c|}{\textbf{BERLIN}} & \multicolumn{3}{c|}{\textbf{ANTWERP}} & \multicolumn{3}{c}{\textbf{MOSCOW}} \\
 & TA-RR$^a$ & MHTE$^b$ & DAF$^c$ & PMD$^d$ & SCL$^e$
 & MAE$\downarrow$ & MSE$\downarrow$ & RMSE$\downarrow$
 & MAE$\downarrow$ & MSE$\downarrow$ & RMSE$\downarrow$
 & MAE$\downarrow$ & MSE$\downarrow$ & RMSE$\downarrow$ \\
\midrule
Baseline & \xmark & \xmark & \xmark & \xmark & \xmark & 0.1247 & 0.8518 & 0.9229 & 0.1189 & 0.8234 & 0.9074 & 0.1356 & 0.8912 & 0.9440 \\
TA-RR only & \cmark & \xmark & \xmark & \xmark & \xmark & 0.1089 & 0.7124 & 0.8440 & 0.1034 & 0.6845 & 0.8274 & 0.1178 & 0.7456 & 0.8635 \\
MHTE only & \xmark & \cmark & \xmark & \xmark & \xmark & 0.1156 & 0.7789 & 0.8826 & 0.1101 & 0.7512 & 0.8667 & 0.1245 & 0.8123 & 0.9013 \\
TA-RR + MHTE & \cmark & \cmark & \xmark & \xmark & \xmark & 0.0952 & 0.6235 & 0.7896 & 0.0901 & 0.5989 & 0.7739 & 0.1023 & 0.6534 & 0.8083 \\
+ DAF & \cmark & \cmark & \cmark & \xmark & \xmark & 0.0831 & 0.5412 & 0.7357 & 0.0789 & 0.5201 & 0.7212 & 0.0901 & 0.5678 & 0.7535 \\
+ PMD & \cmark & \cmark & \cmark & \cmark & \xmark & 0.0756 & 0.4834 & 0.6953 & 0.0712 & 0.4623 & 0.6799 & 0.0823 & 0.5089 & 0.7134 \\
\midrule
\textbf{Full RCSNet} & \cmark & \cmark & \cmark & \cmark & \cmark & \textcolor{red}{\textbf{0.0667}} & \textcolor{red}{\textbf{0.4089}} & \textcolor{red}{\textbf{0.6395}} & \textcolor{red}{\textbf{0.0594}} & \textcolor{red}{\textbf{0.3776}} & \textcolor{red}{\textbf{0.6145}} & \textcolor{red}{\textbf{0.0725}} & \textcolor{red}{\textbf{0.4323}} & \textcolor{red}{\textbf{0.6575}} \\
\bottomrule
\multicolumn{14}{l}{\footnotesize $^a$TA-RR: Topology-Aware Road Representation; $^b$MHTE: Multi-Horizon Temporal Encoder;} \\
\multicolumn{14}{l}{\footnotesize $^c$DAF: Direction-Aware Fusion; $^d$PMD: Progressive Multi-step Decoder; $^e$SCL: Structure-Consistent Learning}
\end{tabular}%
}
\end{table*}

\subsection{Qualitative Analysis}

The qualitative results provide a visual assessment of how well each model preserves the spatial structure and active traffic support of future traffic maps. For each city, three representative time periods are shown to cover different traffic conditions across the day: morning, afternoon, and evening. In each figure, the first three columns show the traffic maps at the selected timestamps, while the final column shows the \textit{aggregated absolute error heatmap}. This heatmap is not the error from only one timestamp. Instead, it is computed by averaging the cell-wise absolute prediction errors across the three selected time periods. Thus, the heatmap summarizes where each model consistently deviates from the ground truth across morning, afternoon, and evening conditions. 

% The color bar represents mean absolute error, where lighter regions indicate smaller prediction errors and darker red regions indicate larger accumulated errors. Since absolute error is used, the heatmap shows error magnitude only and does not cancel over-prediction and under-prediction.

\begin{figure}
    \centering
    \includegraphics[width=1\linewidth]{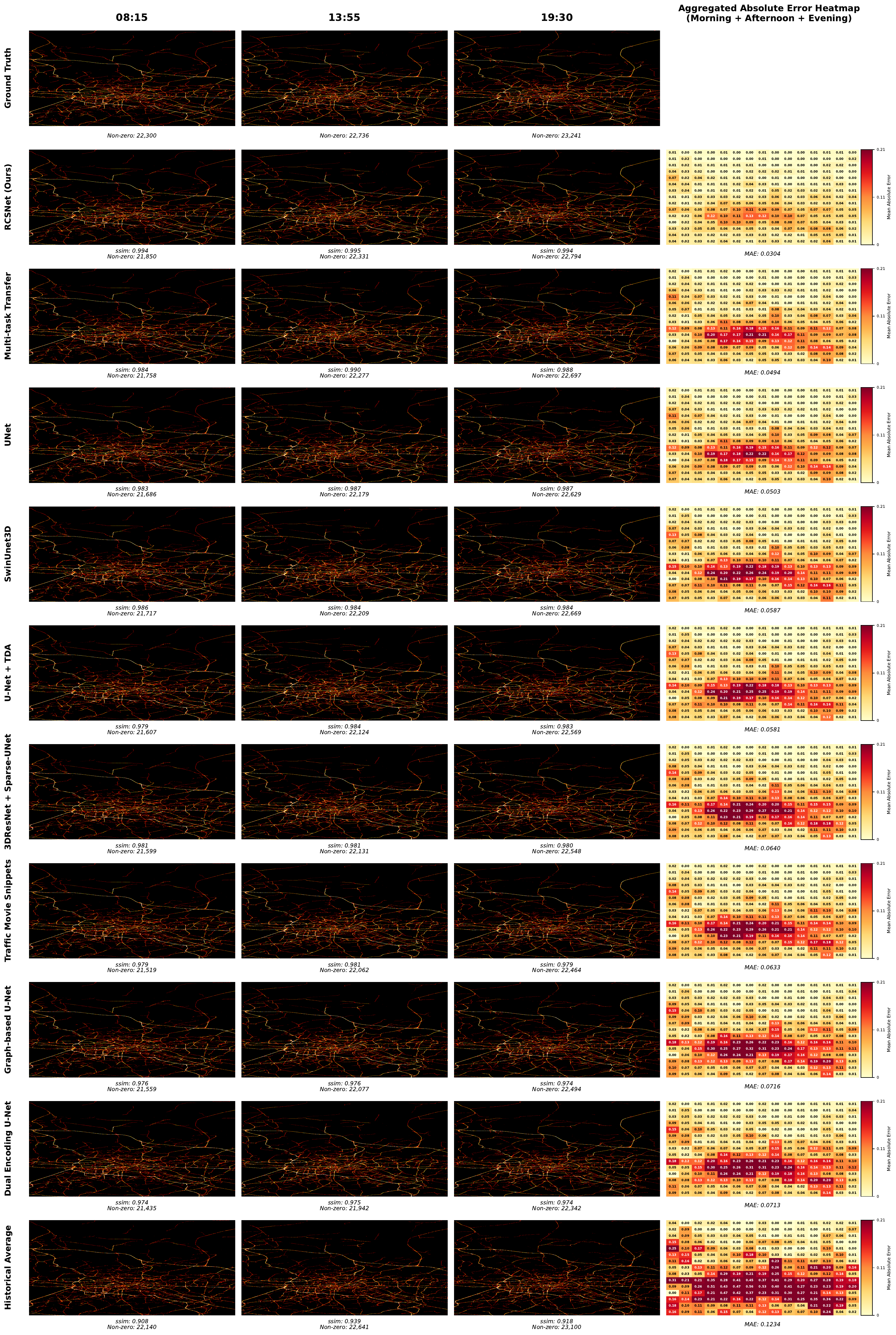}
    \caption{Qualitative comparison of traffic forecasting results in Berlin at 8:15am, 1:55pm, and 7:30pm. RCSNet preserves the main road-aligned traffic structure more closely than the baseline methods, with higher SSIM, lower MAE, and non-zero cell counts closer to the ground truth.}
    \label{fig:berlin}
\end{figure}

In the Berlin example shown in Fig.~\ref{fig:berlin}, RCSNet closely follows the ground-truth traffic structure across all three time periods. The ground-truth maps contain 22,300, 22,736, and 23,241 non-zero cells, while RCSNet predicts 21,850, 22,331, and 22,794 non-zero cells. This close agreement shows that RCSNet preserves the spatial extent of active traffic without excessive spreading or severe underestimation. The model also achieves very high SSIM values of 0.994, 0.995, and 0.994, with an aggregated MAE of 0.0304. The corresponding error heatmap is lighter and more localized than those of the baseline models, indicating stronger spatial agreement with the true traffic distribution. In contrast, several baselines show larger residual errors, while Historical Average gives a higher aggregated MAE of 0.1234 because it cannot adapt well to the selected time-varying traffic states.

\begin{figure}
    \centering
    \includegraphics[width=1\linewidth]{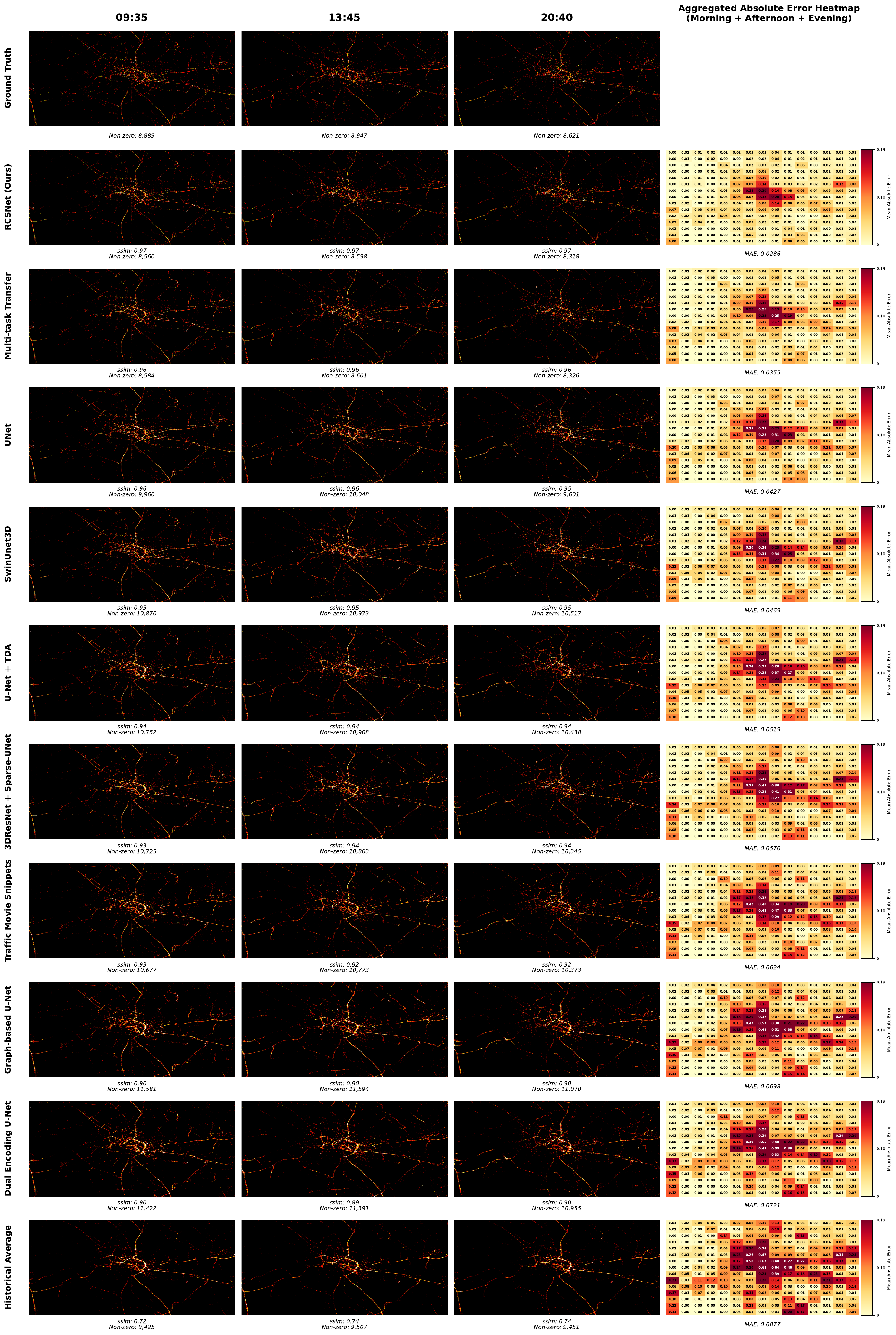}
    \caption{Qualitative comparison of traffic forecasting results in Antwerp at morning, afternoon, and evening periods. RCSNet maintains compact traffic activations that closely match the ground truth, while several baselines produce broader residual errors and less road-aligned predictions.}
    \label{fig:antwerp}
\end{figure}

The Antwerp visualization in Fig.~\ref{fig:antwerp} represents a more compact traffic pattern. The ground-truth maps contain 8,889, 8,947, and 8,621 non-zero cells, while RCSNet predicts 8,560, 8,598, and 8,318 non-zero cells. These values remain close to the true active traffic support, showing that the model maintains the compact road-aligned structure of Antwerp rather than expanding traffic activity into broader regions. The competing models capture the general layout, but their heatmaps show stronger error concentrations, especially near the central road region, and several baselines produce noticeably larger non-zero counts, indicating more spatially spread predictions.

\begin{figure}
    \centering
    \includegraphics[width=1\linewidth]{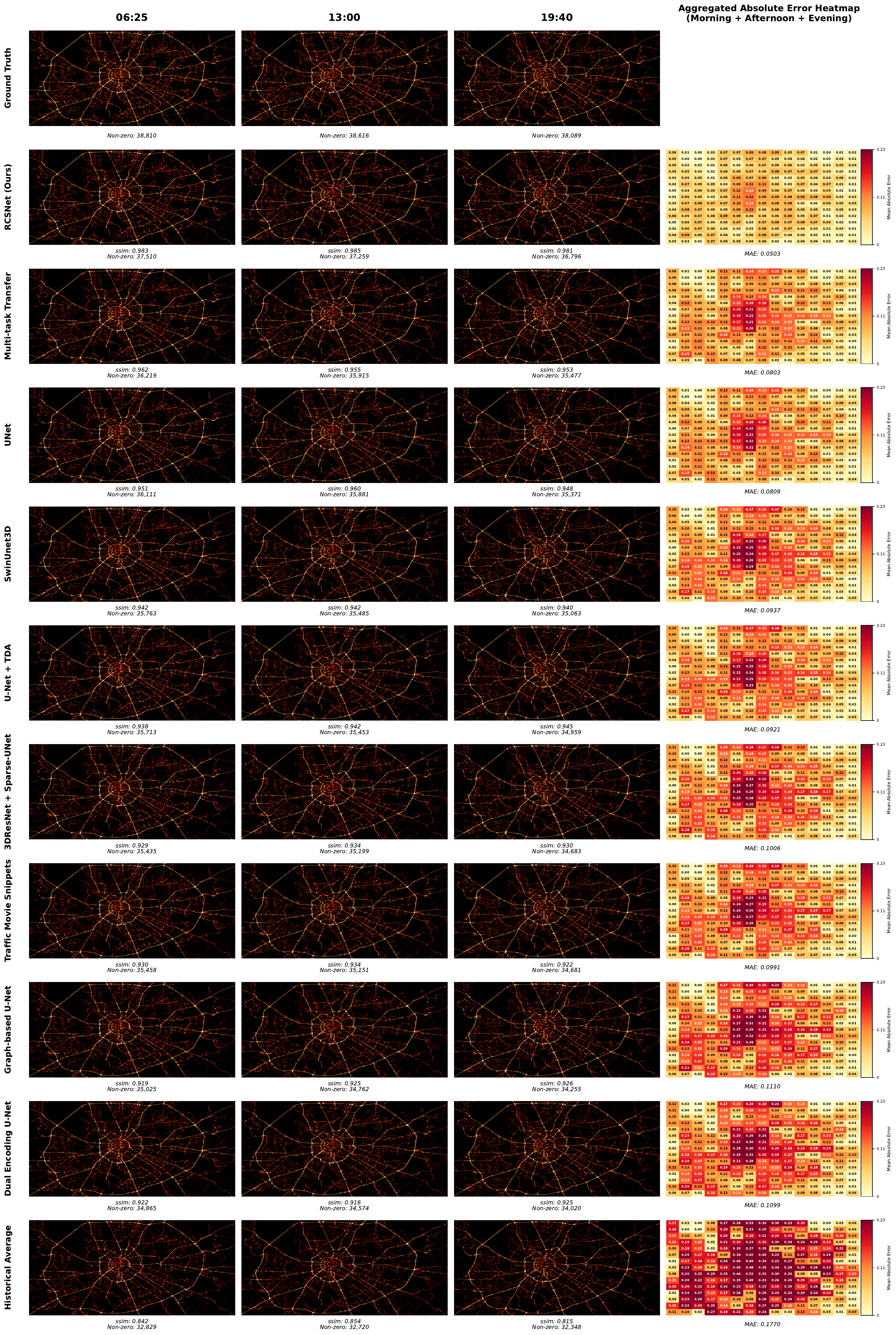}
    \caption{Qualitative comparison of traffic forecasting results in Moscow at morning, afternoon, and evening periods. RCSNet better preserves dense corridor-level traffic patterns and produces weaker difference-map errors than the competing methods.}
    \label{fig:moscow}
\end{figure}

Moscow presents the most challenging visual case in Fig.~\ref{fig:moscow}, mainly because the road network is denser and traffic activity is more widely distributed. The ground-truth maps contain 38,810, 38,616, and 38,089 non-zero cells, while RCSNet predicts 37,510, 37,259, and 36,796 non-zero cells. Although the gap is larger than in Berlin and Antwerp, the predicted traffic support remains closer to the ground truth than the competing methods and preserves the major corridor structure. The larger non-zero counts in Moscow also explain why the visual error appears more pronounced. There are more active road cells, denser intersections, and more distributed traffic patterns to forecast. Even under this more complex setting, RCSNet achieves the best visual agreement, with SSIM values of 0.983, 0.985, and 0.981 and an aggregated MAE of 0.0503. The baseline heatmaps show stronger error accumulation, especially around dense corridors, while RCSNet keeps the errors more controlled.

These qualitative results show that RCSNet produces forecasts that are accurate in value, spatially consistent with the road network, and close to the true active traffic support. The non-zero cell counts confirm that the model does not simply reduce error by suppressing traffic activity; instead, it preserves a realistic amount of active traffic across different city structures. This behavior can be attributed to the topology-aware road representation, multi-horizon temporal encoding, direction-aware fusion, and progressive decoding strategy, which together help the model maintain corridor-level traffic structure in Berlin, compact road-aligned activity in Antwerp, and robust predictions under the dense road layout of Moscow.

% \begin{figure*}
%     \centering
%     \includegraphics[width=0.68\linewidth]{sample_01.pdf}
%     \caption{Enter Caption}
%     \label{fig:qualitative_results}
% \end{figure*}

\subsection{Cross-City Generalization Analysis}

To evaluate transferability to unseen urban environments, a cross-city generalization experiment is conducted using Berlin, Antwerp, and Moscow as source cities, and Chicago and Bangkok as unseen target cities. During testing, each model receives recent historical traffic frames from the target city and predicts the corresponding future frames. For a fair comparison, the static road map is also provided to all learnable models as structural input. The baseline models use the road map through input-level concatenation where applicable, while RCSNet explicitly encodes and fuses the road map through its topology-aware road representation and direction-aware fusion modules.

As shown in Table~\ref{tab:cross_city}, RCSNet achieves the best performance on both unseen cities across all evaluation metrics. On Chicago, RCSNet obtains an MAE of 0.1312, MSE of 0.4986, and RMSE of 0.7061. Compared with the closest competing baseline, Multi-task Transfer Learning, this corresponds to reductions of 13.9\% in MAE, 20.1\% in MSE, and 10.6\% in RMSE. This shows that RCSNet transfers more effectively to the unseen Chicago traffic structure, even when all learnable models are provided with the same static road information.

A similar trend is observed on Bangkok. RCSNet achieves an MAE of 0.1423, MSE of 0.5407, and RMSE of 0.7353, outperforming all baseline methods. Relative to Multi-task Transfer Learning, RCSNet reduces MAE by 13.5\%, MSE by 19.8\%, and RMSE by 10.5\%. These consistent gains across two unseen target cities indicate that RCSNet does not simply memorize source-city traffic patterns. Instead, it learns a more transferable relationship between recent traffic observations and the road structure that supports traffic movement.

\begin{figure}
    \centering
    \includegraphics[width=1\linewidth]{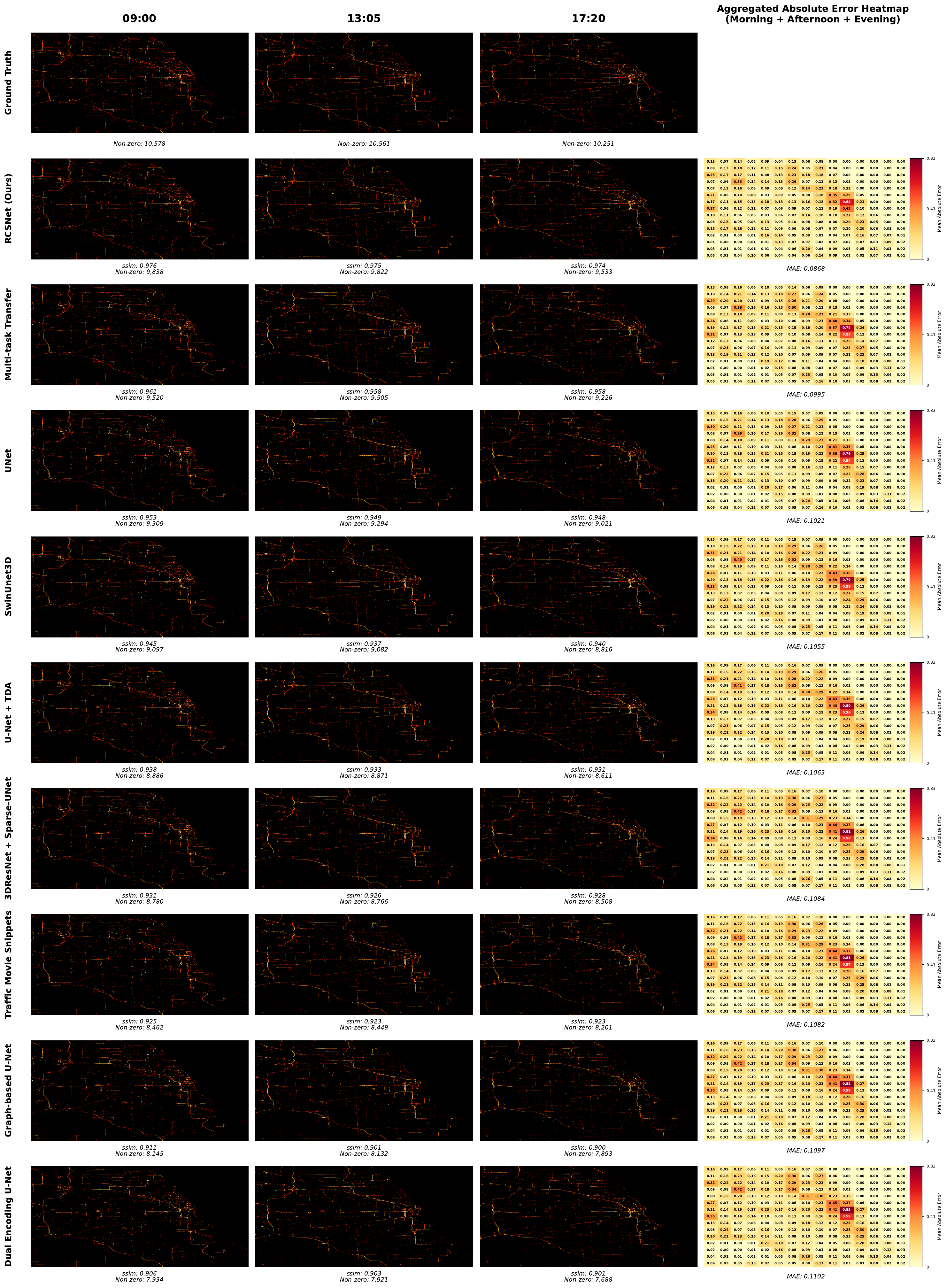}
    \caption{Cross-city qualitative forecasting results on Chicago. The first three columns show morning, afternoon, and evening predictions, while the final column shows the aggregated absolute error heatmap across the selected times.}
    \label{fig:chicago}
\end{figure}

\begin{figure}
    \centering
    \includegraphics[width=1\linewidth]{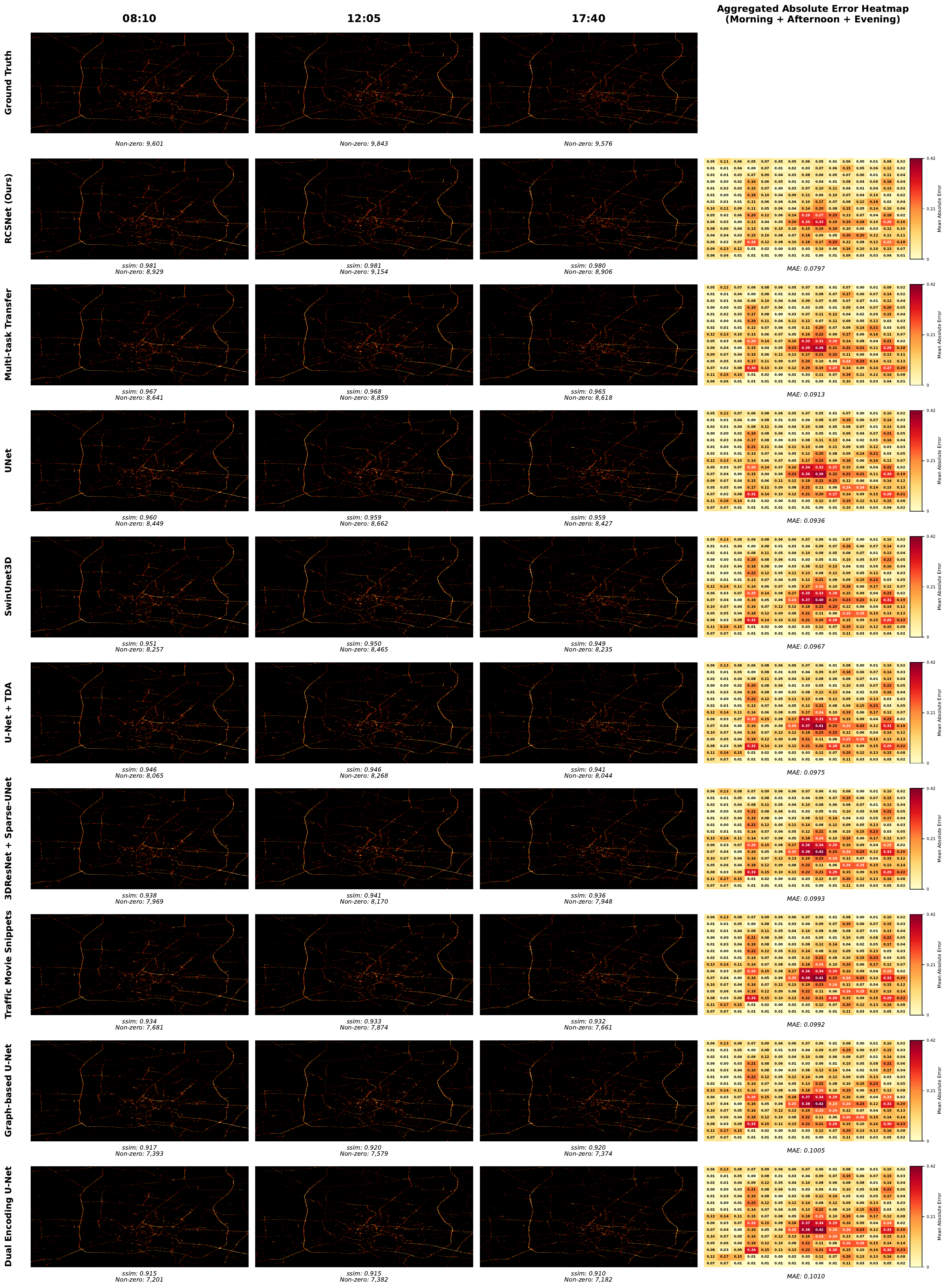}
    \caption{Cross-city qualitative forecasting results on Bangkok.  The first three columns show morning, afternoon, and evening predictions, while the final column reports the aggregated absolute error heatmap across the selected times.}
    \label{fig:bangkok}
\end{figure}

The qualitative cross-city results (as illustrated in Fig~\ref{fig:chicago} and Fig~\ref{fig:bangkok}) further support this finding. In Chicago, RCSNet maintains SSIM values of 0.981, 0.981, and 0.980 across the selected morning, afternoon, and evening examples, with an aggregated MAE of 0.0797. In Bangkok, it achieves SSIM values of 0.976, 0.975, and 0.974, with an aggregated MAE of 0.0868. The aggregated absolute error heatmaps show that RCSNet produces smaller and more spatially contained error regions than the baseline methods. This means that the remaining errors are less widely spread across the traffic map and are more confined to specific active regions, indicating stronger spatial consistency under unseen-city conditions.

These results demonstrate that the proposed road-conditioned architecture improves cross-city generalization beyond simply providing road-map information as an input. While the baselines receive the static map as an additional channel, RCSNet uses road topology more explicitly to guide spatial representation, road-traffic interaction, and future-state decoding. This allows the model to produce future traffic maps that are both more accurate and more consistent with unseen urban road structures.

\begin{table*}[!t]
\centering
\scriptsize
\caption{Cross-city generalization performance. Models trained on BERLIN, ANTWERP, and MOSCOW, tested on unseen cities (CHICAGO, BANGKOK). Best results in \textbf{\textcolor{red}{red}}, second best \underline{underlined}.}
\label{tab:cross_city}
\renewcommand{\arraystretch}{1.10}
\resizebox{\textwidth}{!}{%
\begin{tabular}{l | r r r | r r r}
\toprule
\multirow{2}{*}{\textbf{Method}} & \multicolumn{3}{c|}{\textbf{CHICAGO}} & \multicolumn{3}{c}{\textbf{BANGKOK}} \\
 & MAE$\downarrow$ & MSE$\downarrow$ & RMSE$\downarrow$ & MAE$\downarrow$ & MSE$\downarrow$ & RMSE$\downarrow$ \\
\midrule
Multi-task Transfer Learning \cite{lu2021learning}  & \underline{0.1523} & \underline{0.6244} & \underline{0.7902} & \underline{0.1645} & \underline{0.6744} & \underline{0.8212} \\
UNet \cite{choi2020utilizing}                        & 0.1589 & 0.6833 & 0.8266 & 0.1712 & 0.7362 & 0.8580 \\
SwinUnet3D \cite{bojesomo2022swinunet3d}             & 0.1634 & 0.7353 & 0.8575 & 0.1778 & 0.8001 & 0.8945 \\
U-Net + TDA \cite{konyakhin2021solving}              & 0.1678 & 0.7719 & 0.8786 & 0.1823 & 0.8386 & 0.9158 \\
Traffic Movie Snippets \cite{wiedemann2021traffic}   & 0.1712 & 0.8218 & 0.9065 & 0.1856 & 0.8909 & 0.9439 \\
3DResNet + Sparse-UNet \cite{wang2021traffic4cast}   & 0.1723 & 0.8270 & 0.9094 & 0.1867 & 0.8962 & 0.9467 \\
Graph-based U-Net \cite{hermes2022graph}             & 0.1798 & 0.8990 & 0.9482 & 0.1934 & 0.9670 & 0.9834 \\
Dual Encoding U-Net \cite{santokhi2021dual}          & 0.1834 & 0.9353 & 0.9671 & 0.1978 & 1.0088 & 1.0044 \\
\midrule
\textbf{\textcolor{red}{RCSNet (Ours)}} & \textbf{\textcolor{red}{0.1312}} & \textbf{\textcolor{red}{0.4986}} & \textbf{\textcolor{red}{0.7061}} & \textbf{\textcolor{red}{0.1423}} & \textbf{\textcolor{red}{0.5407}} & \textbf{\textcolor{red}{0.7353}} \\
\bottomrule
\end{tabular}%
}
\end{table*}

% \begin{figure}
%     \centering
%     \includegraphics[width=1\linewidth]{Chicago_Cross_City.pdf}
%     \caption{Enter Caption}
%     \label{fig:placeholder}
% \end{figure}

\subsection{Forecast Horizon Analysis}

\begin{figure}
    \centering
    \includegraphics[width=1\linewidth]{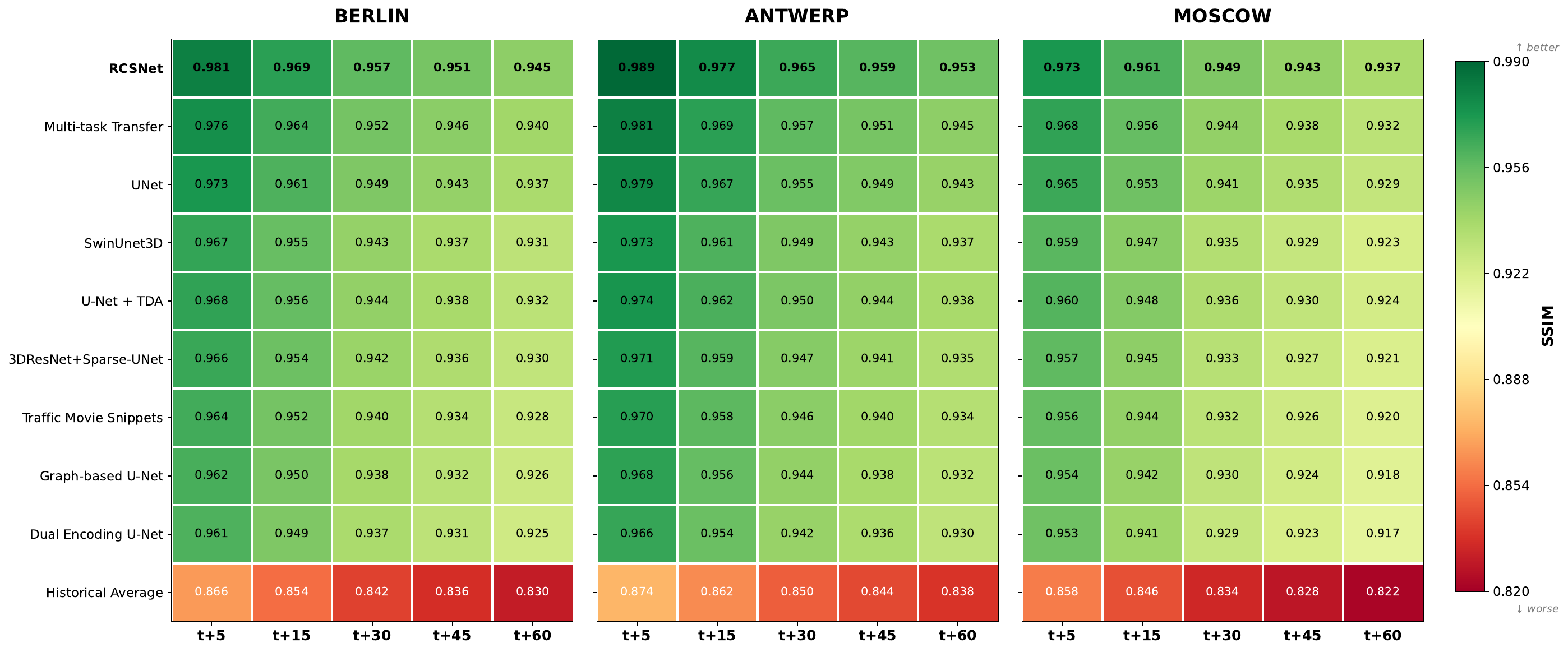}
    \caption{Forecast horizon SSIM heatmap across Berlin, Antwerp, and Moscow. RCSNet maintains the strongest performance across short- and long-term horizons, showing slower degradation from $t+5$ to $t+60$ minutes compared with the baseline methods.}

    \label{fig:horizon_heatmap}
\end{figure}

To further evaluate the temporal stability of the proposed model, forecast horizon analysis is conducted over five future horizons, namely $t+5$, $t+15$, $t+30$, $t+45$, and $t+60$ minutes. Since the model predicts the next 12 traffic frames, these horizons represent short-, medium-, and longer-term forecasting behavior within the one-hour prediction window. Table~\ref{tab:horizon_analysis} reports MAE, MSE, and RMSE for each horizon across Berlin, Antwerp, and Moscow, while Fig.~\ref{fig:horizon_heatmap} provides a visual comparison of horizon-wise performance trends.

The results show that forecasting error increases as the prediction horizon becomes longer, which is expected because future traffic states become more uncertain over time. This trend is observed for all methods and all cities. However, RCSNet consistently achieves the lowest error across every horizon. In Berlin, RCSNet reduces RMSE from 0.5350 at $t+5$ to 0.6395 at $t+60$, while the closest baseline, Multi-task Transfer Learning, increases from 0.5620 to 0.6717. A similar pattern is observed in Antwerp, where RCSNet obtains the lowest RMSE at all horizons, increasing from 0.5141 at $t+5$ to 0.6145 at $t+60$. In Moscow, which has a denser and more complex traffic structure, RCSNet also remains the best-performing method, with RMSE increasing from 0.5501 at $t+5$ to 0.6575 at $t+60$.

The SSIM heatmap in Fig.~\ref{fig:horizon_heatmap} further confirms this pattern by showing that RCSNet maintains the strongest horizon-wise scores across the three cities. Although all models experience gradual performance degradation from short-term to longer-term prediction, RCSNet shows a slower decline than the competing methods. This indicates that the proposed model is better able to preserve temporal consistency as the forecast horizon increases. The improvement is especially important at longer horizons, where small errors from earlier predictions can accumulate and lead to unstable future traffic maps.

These results demonstrate the benefit of the progressive multi-step decoder and the multi-horizon temporal encoder. The multi-horizon encoder captures short-term fluctuations as well as longer traffic trends, while the progressive decoder generates future frames sequentially rather than treating each horizon as an isolated prediction. Together with the road-conditioned fusion and structure-consistent learning objective, this design helps RCSNet maintain more stable and road-aligned forecasts across the full prediction horizon.

\begin{table*}[!t]
\centering
\scriptsize
\caption{Forecast horizon performance across multiple cities. Performance at t+5, t+15, t+30, t+45, and t+60 minute horizons. Best results per horizon in \textcolor{red}{Red}, second best \underline{underlined}.}
\label{tab:horizon_analysis}
\renewcommand{\arraystretch}{1.15}
\resizebox{\textwidth}{!}{%
\begin{tabular}{l l | c c c | c c c | c c c | c c c | c c c}
\toprule
\multirow{2}{*}{\textbf{City}} & \multirow{2}{*}{\textbf{Method}} & \multicolumn{3}{c|}{\textbf{t+5 min}} & \multicolumn{3}{c|}{\textbf{t+15 min}} & \multicolumn{3}{c|}{\textbf{t+30 min}} & \multicolumn{3}{c|}{\textbf{t+45 min}} & \multicolumn{3}{c}{\textbf{t+60 min}} \\
 & & MAE & MSE & RMSE & MAE & MSE & RMSE & MAE & MSE & RMSE & MAE & MSE & RMSE & MAE & MSE & RMSE \\
\midrule
\multirow{10}{*}{BERLIN}
 & Historical Average          & 0.1627 & 0.8108 & 0.9004 & 0.1859 & 0.9266 & 0.9626 & 0.2092 & 1.0425 & 1.0210 & 0.2208 & 1.1004 & 1.0490 & 0.2324 & 1.1583 & 1.0762 \\
 & Multi-task Transfer \cite{lu2021learning}  & \underline{0.0521} & \underline{0.3158} & \underline{0.5620} & \underline{0.0596} & \underline{0.3610} & \underline{0.6008} & \underline{0.0670} & \underline{0.4061} & \underline{0.6373} & \underline{0.0708} & \underline{0.4286} & \underline{0.6547} & \underline{0.0745} & \underline{0.4512} & \underline{0.6717} \\
 & UNet \cite{choi2020utilizing}              & 0.0547 & 0.3282 & 0.5729 & 0.0626 & 0.3751 & 0.6125 & 0.0704 & 0.4220 & 0.6496 & 0.0743 & 0.4455 & 0.6675 & 0.0782 & 0.4689 & 0.6848 \\
 & U-Net + TDA \cite{konyakhin2021solving}    & 0.0599 & 0.3664 & 0.6053 & 0.0685 & 0.4187 & 0.6471 & 0.0770 & 0.4711 & 0.6864 & 0.0813 & 0.4972 & 0.7051 & 0.0856 & 0.5234 & 0.7235 \\
 & Traffic Movie \cite{wiedemann2021traffic}  & 0.0638 & 0.3897 & 0.6243 & 0.0730 & 0.4454 & 0.6674 & 0.0821 & 0.5010 & 0.7078 & 0.0866 & 0.5289 & 0.7273 & 0.0912 & 0.5567 & 0.7461 \\
 & 3DResNet \cite{wang2021traffic4cast}       & 0.0622 & 0.3781 & 0.6149 & 0.0711 & 0.4321 & 0.6573 & 0.0800 & 0.4861 & 0.6972 & 0.0845 & 0.5131 & 0.7163 & 0.0889 & 0.5401 & 0.7349 \\
 & Graph U-Net \cite{hermes2022graph}         & 0.0654 & 0.3975 & 0.6305 & 0.0747 & 0.4542 & 0.6739 & 0.0841 & 0.5110 & 0.7148 & 0.0887 & 0.5394 & 0.7344 & 0.0934 & 0.5678 & 0.7535 \\
 & SwinUnet3D \cite{bojesomo2022swinunet3d}   & 0.0607 & 0.3702 & 0.6084 & 0.0694 & 0.4231 & 0.6505 & 0.0780 & 0.4760 & 0.6899 & 0.0824 & 0.5025 & 0.7089 & 0.0867 & 0.5289 & 0.7273 \\
 & Dual Encoding \cite{santokhi2021dual}      & 0.0669 & 0.4052 & 0.6366 & 0.0765 & 0.4631 & 0.6805 & 0.0860 & 0.5210 & 0.7218 & 0.0908 & 0.5500 & 0.7416 & 0.0956 & 0.5789 & 0.7609 \\
 & \textbf{RCSNet}             & \textcolor{red}{\textbf{0.0467}} & \textcolor{red}{\textbf{0.2862}} & \textcolor{red}{\textbf{0.5350}} & \textcolor{red}{\textbf{0.0534}} & \textcolor{red}{\textbf{0.3271}} & \textcolor{red}{\textbf{0.5719}} & \textcolor{red}{\textbf{0.0600}} & \textcolor{red}{\textbf{0.3680}} & \textcolor{red}{\textbf{0.6066}} & \textcolor{red}{\textbf{0.0634}} & \textcolor{red}{\textbf{0.3885}} & \textcolor{red}{\textbf{0.6233}} & \textcolor{red}{\textbf{0.0667}} & \textcolor{red}{\textbf{0.4089}} & \textcolor{red}{\textbf{0.6395}} \\
\midrule
\multirow{10}{*}{ANTWERP}
 & Historical Average          & 0.1509 & 0.7626 & 0.8733 & 0.1725 & 0.8715 & 0.9335 & 0.1940 & 0.9805 & 0.9902 & 0.2048 & 1.0349 & 1.0173 & 0.2156 & 1.0894 & 1.0437 \\
 & Multi-task Transfer \cite{lu2021learning}  & \underline{0.0489} & \underline{0.2964} & \underline{0.5444} & \underline{0.0558} & \underline{0.3387} & \underline{0.5820} & \underline{0.0628} & \underline{0.3811} & \underline{0.6173} & \underline{0.0663} & \underline{0.4022} & \underline{0.6342} & \underline{0.0698} & \underline{0.4234} & \underline{0.6507} \\
 & UNet \cite{choi2020utilizing}              & 0.0514 & 0.3081 & 0.5551 & 0.0587 & 0.3521 & 0.5934 & 0.0661 & 0.3961 & 0.6294 & 0.0697 & 0.4181 & 0.6466 & 0.0734 & 0.4401 & 0.6634 \\
 & U-Net + TDA \cite{konyakhin2021solving}    & 0.0561 & 0.3438 & 0.5863 & 0.0641 & 0.3930 & 0.6269 & 0.0721 & 0.4421 & 0.6649 & 0.0761 & 0.4666 & 0.6831 & 0.0801 & 0.4912 & 0.7009 \\
 & Traffic Movie \cite{wiedemann2021traffic}  & 0.0599 & 0.3664 & 0.6053 & 0.0685 & 0.4187 & 0.6471 & 0.0770 & 0.4711 & 0.6864 & 0.0813 & 0.4972 & 0.7051 & 0.0856 & 0.5234 & 0.7235 \\
 & 3DResNet \cite{wang2021traffic4cast}       & 0.0584 & 0.3547 & 0.5956 & 0.0667 & 0.4054 & 0.6367 & 0.0751 & 0.4560 & 0.6753 & 0.0792 & 0.4814 & 0.6938 & 0.0834 & 0.5067 & 0.7118 \\
 & Graph U-Net \cite{hermes2022graph}         & 0.0615 & 0.3734 & 0.6111 & 0.0702 & 0.4267 & 0.6532 & 0.0790 & 0.4801 & 0.6929 & 0.0834 & 0.5067 & 0.7118 & 0.0878 & 0.5334 & 0.7303 \\
 & SwinUnet3D \cite{bojesomo2022swinunet3d}   & 0.0568 & 0.3477 & 0.5897 & 0.0650 & 0.3974 & 0.6304 & 0.0731 & 0.4470 & 0.6686 & 0.0771 & 0.4719 & 0.6869 & 0.0812 & 0.4967 & 0.7048 \\
 & Dual Encoding \cite{santokhi2021dual}      & 0.0631 & 0.3811 & 0.6173 & 0.0721 & 0.4356 & 0.6600 & 0.0811 & 0.4900 & 0.7000 & 0.0856 & 0.5173 & 0.7192 & 0.0901 & 0.5445 & 0.7379 \\
 & \textbf{RCSNet}             & \textcolor{red}{\textbf{0.0416}} & \textcolor{red}{\textbf{0.2643}} & \textcolor{red}{\textbf{0.5141}} & \textcolor{red}{\textbf{0.0475}} & \textcolor{red}{\textbf{0.3021}} & \textcolor{red}{\textbf{0.5496}} & \textcolor{red}{\textbf{0.0535}} & \textcolor{red}{\textbf{0.3398}} & \textcolor{red}{\textbf{0.5829}} & \textcolor{red}{\textbf{0.0564}} & \textcolor{red}{\textbf{0.3587}} & \textcolor{red}{\textbf{0.5989}} & \textcolor{red}{\textbf{0.0594}} & \textcolor{red}{\textbf{0.3776}} & \textcolor{red}{\textbf{0.6145}} \\
\midrule
\multirow{10}{*}{MOSCOW}
 & Historical Average          & 0.1742 & 0.8517 & 0.9229 & 0.1991 & 0.9734 & 0.9866 & 0.2240 & 1.0950 & 1.0464 & 0.2365 & 1.1559 & 1.0751 & 0.2489 & 1.2167 & 1.1030 \\
 & Multi-task Transfer \cite{lu2021learning}  & \underline{0.0561} & \underline{0.3352} & \underline{0.5790} & \underline{0.0641} & \underline{0.3831} & \underline{0.6190} & \underline{0.0721} & \underline{0.4310} & \underline{0.6565} & \underline{0.0761} & \underline{0.4550} & \underline{0.6745} & \underline{0.0801} & \underline{0.4789} & \underline{0.6920} \\
 & UNet \cite{choi2020utilizing}              & 0.0592 & 0.3469 & 0.5890 & 0.0676 & 0.3965 & 0.6297 & 0.0761 & 0.4460 & 0.6678 & 0.0803 & 0.4708 & 0.6861 & 0.0845 & 0.4956 & 0.7040 \\
 & U-Net + TDA \cite{konyakhin2021solving}    & 0.0646 & 0.3842 & 0.6198 & 0.0738 & 0.4391 & 0.6626 & 0.0831 & 0.4940 & 0.7029 & 0.0877 & 0.5215 & 0.7221 & 0.0923 & 0.5489 & 0.7409 \\
 & Traffic Movie \cite{wiedemann2021traffic}  & 0.0692 & 0.4131 & 0.6427 & 0.0791 & 0.4721 & 0.6871 & 0.0890 & 0.5311 & 0.7288 & 0.0940 & 0.5606 & 0.7487 & 0.0989 & 0.5901 & 0.7682 \\
 & 3DResNet \cite{wang2021traffic4cast}       & 0.0677 & 0.4014 & 0.6336 & 0.0774 & 0.4587 & 0.6773 & 0.0870 & 0.5161 & 0.7184 & 0.0919 & 0.5447 & 0.7380 & 0.0967 & 0.5734 & 0.7572 \\
 & Graph U-Net \cite{hermes2022graph}         & 0.0708 & 0.4208 & 0.6487 & 0.0810 & 0.4810 & 0.6935 & 0.0911 & 0.5411 & 0.7356 & 0.0961 & 0.5711 & 0.7557 & 0.1012 & 0.6012 & 0.7754 \\
 & SwinUnet3D \cite{bojesomo2022swinunet3d}   & 0.0662 & 0.3936 & 0.6274 & 0.0756 & 0.4498 & 0.6707 & 0.0851 & 0.5061 & 0.7114 & 0.0898 & 0.5342 & 0.7309 & 0.0945 & 0.5623 & 0.7499 \\
 & Dual Encoding \cite{santokhi2021dual}      & 0.0724 & 0.4294 & 0.6553 & 0.0827 & 0.4907 & 0.7005 & 0.0931 & 0.5521 & 0.7430 & 0.0982 & 0.5827 & 0.7633 & 0.1034 & 0.6134 & 0.7832 \\
 & \textbf{RCSNet}             & \textcolor{red}{\textbf{0.0507}} & \textcolor{red}{\textbf{0.3026}} & \textcolor{red}{\textbf{0.5501}} & \textcolor{red}{\textbf{0.0580}} & \textcolor{red}{\textbf{0.3458}} & \textcolor{red}{\textbf{0.5880}} & \textcolor{red}{\textbf{0.0653}} & \textcolor{red}{\textbf{0.3891}} & \textcolor{red}{\textbf{0.6238}} & \textcolor{red}{\textbf{0.0689}} & \textcolor{red}{\textbf{0.4107}} & \textcolor{red}{\textbf{0.6409}} & \textcolor{red}{\textbf{0.0725}} & \textcolor{red}{\textbf{0.4323}} & \textcolor{red}{\textbf{0.6575}} \\
\bottomrule
\end{tabular}%
}
\end{table*}

\subsection{Road-Structure Consistency Analysis}

To further examine whether the predicted traffic maps respect the physical road network, a road-structure consistency analysis is conducted using the static road mask of each city. Unlike the standard MAE, which averages prediction error over the full spatial grid, this analysis evaluates whether the model predicts traffic accurately on valid road regions and avoids unrealistic traffic activations outside the road network. Three metrics are used: road MAE, off-road activation rate, and road coverage recall.

\begin{figure}
    \centering
    \includegraphics[width=1\linewidth]{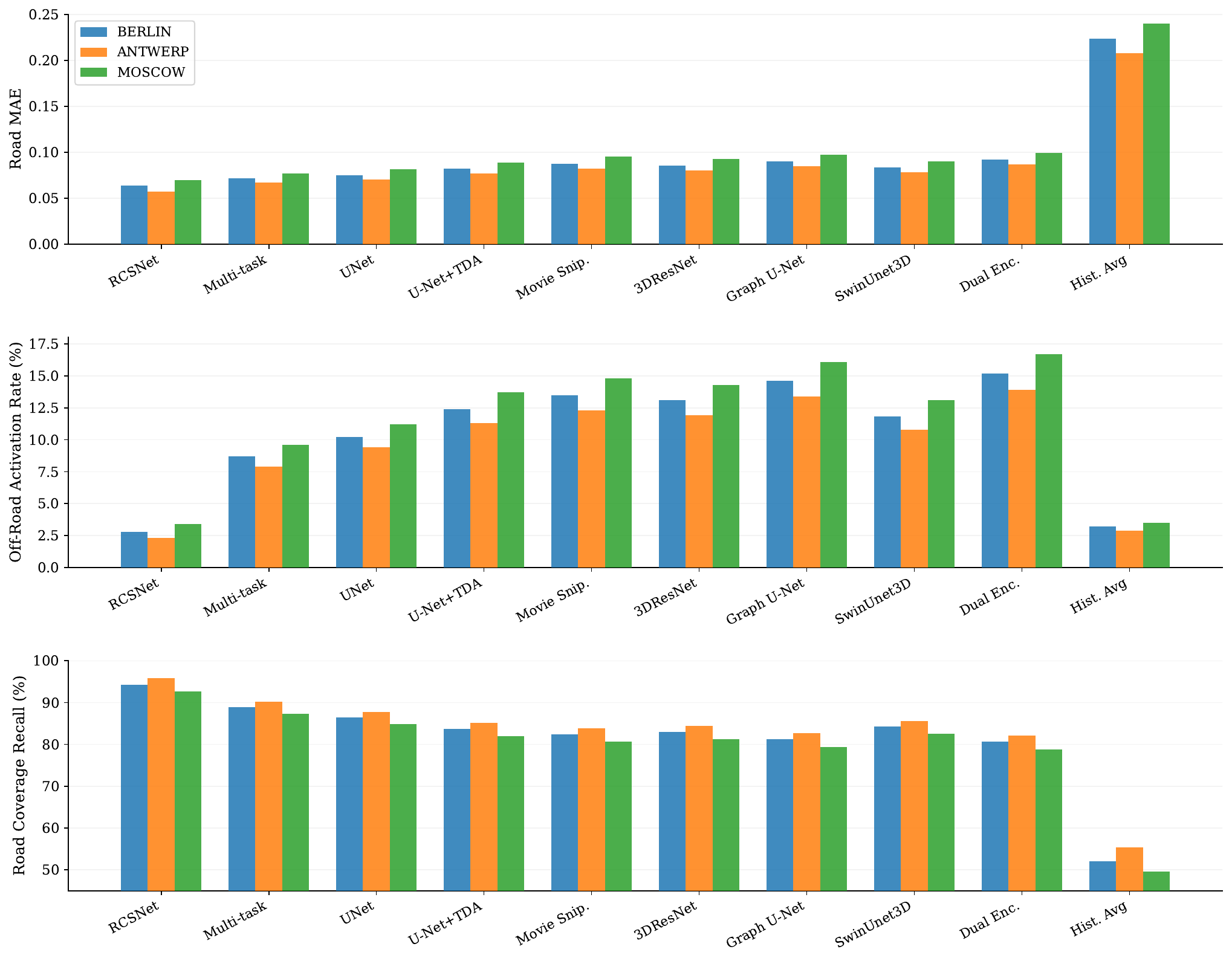}
    \caption{Road-structure consistency comparison across Berlin, Antwerp, and Moscow. RCSNet achieves lower road-region error, fewer off-road activations, and higher road coverage recall, indicating stronger alignment between predicted traffic maps and the underlying road network.}
    \label{fig:road_structure_consistency}
\end{figure}

As shown in Fig.~\ref{fig:road_structure_consistency}, RCSNet achieves the lowest road MAE across Berlin, Antwerp, and Moscow. Compared with the closest competing baseline, Multi-task Transfer Learning, RCSNet reduces road MAE by 10.6\% in Berlin, from 0.0718 to 0.0642; by 15.0\% in Antwerp, from 0.0673 to 0.0572; and by 9.6\% in Moscow, from 0.0772 to 0.0698. These reductions show that RCSNet improves prediction accuracy specifically on road regions, where traffic activity is physically meaningful.

RCSNet also achieves the lowest off-road activation rate among the learned models. Relative to Multi-task Transfer Learning, it reduces off-road activation by 67.8\% in Berlin, from 8.7\% to 2.8\%; by 70.9\% in Antwerp, from 7.9\% to 2.3\%; and by 64.6\% in Moscow, from 9.6\% to 3.4\%. This indicates that RCSNet is much less likely to spread predicted traffic into non-road regions. Although Historical Average has a similarly low off-road activation rate, this should not be interpreted as better structural consistency. Its low off-road activation mainly results from conservative predictions that suppress traffic activity overall.

The road coverage recall results clarify this distinction. RCSNet achieves recall values of 94.3\%, 95.8\%, and 92.7\% in Berlin, Antwerp, and Moscow, respectively. Compared with Multi-task Transfer Learning, this corresponds to gains of +5.4, +5.6, and +5.4 percentage points. In contrast, Historical Average reaches only 52.1\%, 55.3\%, and 49.6\% recall in the three cities, meaning that RCSNet improves road coverage recall over Historical Average by +42.2, +40.5, and +43.1 percentage points. This confirms that Historical Average avoids off-road activations mainly by missing many valid road-region traffic cells, while RCSNet maintains both high coverage and strong spatial precision.

These results show that RCSNet provides a better balance between road-region accuracy, off-road suppression, and traffic coverage. The model reduces errors on valid road cells, limits false activations outside the road network, and still recovers a larger portion of true road traffic activity. This behavior supports the role of the topology-aware road representation, direction-aware fusion, and structure-consistent learning objective in producing forecasts that are both accurate and physically aligned with the road network.

\subsection{Statistical Analysis}

A statistical analysis was conducted to further evaluate the consistency and reliability of RCSNet across the three evaluated cities. Table~\ref{tab:mean_performance_cities} summarizes the mean performance and standard deviations of each method across Berlin, Antwerp, and Moscow. From Table~\ref{tab:mean_performance_cities}, it is evident that RCSNet achieves the best average results across all evaluation metrics. Specifically, the proposed model obtains the lowest MAE ($0.0662 \pm 0.0067$), MSE ($0.4063 \pm 0.0291$), and RMSE ($0.6349 \pm 0.0243$), indicating more accurate and stable forecasting performance across different urban traffic environments. Compared with the closest competing baseline, Multi-task Transfer Learning, RCSNet reduces the average MAE from 0.0748 to 0.0662, the MSE from 0.4512 to 0.4063, and the RMSE from 0.6715 to 0.6349. These correspond to reductions of 11.5\%, 10.0\%, and 5.5\%, respectively. The lower standard deviations reported by RCSNet also suggest that its performance is less sensitive to city-specific differences, which further supports its robustness across varying road layouts and traffic patterns.

To determine whether these improvements are statistically meaningful, paired $t$-tests were conducted between RCSNet and each baseline method, as shown in Table~\ref{tab:statistical_significance}. The paired test is appropriate because all methods were evaluated on the same three cities, allowing the city-level differences in performance to be compared directly. Positive mean differences indicate that RCSNet achieves lower error than the corresponding baseline. From Table~\ref{tab:statistical_significance}, RCSNet shows statistically significant improvements in both MAE and RMSE compared with all baseline methods. For MAE, the proposed model significantly outperforms Multi-task Transfer Learning and UNet, with mean differences of $+0.0086$ and $+0.0125$, respectively, both significant at $p<0.05$. Stronger significance is observed against SwinUnet3D, U-Net with TDA, 3DResNet, Traffic Movie Snippets, Graph-based U-Net, Dual Encoding U-Net, and Historical Average, where the $p$-values fall below $0.01$ or $0.001$.

A similar trend is observed for RMSE. RCSNet achieves significant improvements over Multi-task Transfer Learning and UNet, with mean differences of $+0.0366$ and $+0.0492$, respectively. The improvements are larger against the remaining baselines, especially Dual Encoding U-Net, Graph-based U-Net, Traffic Movie Snippets, and Historical Average. These results confirm that the performance gains of RCSNet are not only numerically consistent across cities, but also provides supporting evidence.

\begin{table*}[t]
\centering
\caption{Mean performance and standard deviation by model across all cities (BERLIN, ANTWERP, MOSCOW). Lower values indicate better performance. Best results in \textbf{bold}.}
\label{tab:mean_performance_cities}
\renewcommand{\arraystretch}{1.15}
\small
\begin{tabular}{l c c c}
\toprule
\textbf{Model} & \textbf{MAE} $\downarrow$ & \textbf{MSE} $\downarrow$ & \textbf{RMSE} $\downarrow$ \\
\midrule
Historical Average              & 0.2323 $\pm$ 0.0314 & 1.5815 $\pm$ 0.5137 & 0.8543 $\pm$ 0.1823 \\
{Multi-task Transfer Learning} & {0.0748 $\pm$ 0.0086} & {0.4512 $\pm$ 0.0431} & {0.6715 $\pm$ 0.0318} \\
UNet                            & 0.0787 $\pm$ 0.0097 & 0.4682 $\pm$ 0.0514 & 0.6841 $\pm$ 0.0362 \\
U-Net + TDA                     & 0.0860 $\pm$ 0.0113 & 0.5212 $\pm$ 0.0571 & 0.7217 $\pm$ 0.0403 \\
Traffic Movie Snippets          & 0.0919 $\pm$ 0.0151 & 0.5567 $\pm$ 0.0731 & 0.7459 $\pm$ 0.0483 \\
3DResNet + Sparse-UNet          & 0.0897 $\pm$ 0.0138 & 0.5401 $\pm$ 0.0682 & 0.7346 $\pm$ 0.0461 \\
Graph-based U-Net               & 0.0941 $\pm$ 0.0172 & 0.5675 $\pm$ 0.0793 & 0.7531 $\pm$ 0.0514 \\
SwinUnet3D                      & 0.0875 $\pm$ 0.0126 & 0.5293 $\pm$ 0.0623 & 0.7273 $\pm$ 0.0432 \\
Dual Encoding U-Net             & 0.0964 $\pm$ 0.0193 & 0.5789 $\pm$ 0.0874 & 0.7607 $\pm$ 0.0561 \\
\midrule
\textbf{RCSNet (Ours)}          & \textbf{0.0662 $\pm$ 0.0067} & \textbf{0.4063 $\pm$ 0.0291} & \textbf{0.6349 $\pm$ 0.0243} \\
\bottomrule
\end{tabular}
\end{table*}

% \begin{table}[t]
% \centering
% \caption{Mean performance and Standard deviations across source cities. Lower values are better; best results are in \textbf{bold}.}
% \label{tab:mean_performance_cities}
% \renewcommand{\arraystretch}{0.95}
% \scriptsize
% \setlength{\tabcolsep}{2.5pt}
% \resizebox{\columnwidth}{!}{%
% \begin{tabular}{lccc}
% \toprule
% \textbf{Model} & \textbf{MAE}$\downarrow$ & \textbf{MSE}$\downarrow$ & \textbf{RMSE}$\downarrow$ \\
% \midrule
% Hist. Avg.      & $0.2323{\pm}0.0314$ & $1.5815{\pm}0.5137$ & $0.8543{\pm}0.1823$ \\
% MT-Transfer     & $0.0748{\pm}0.0086$ & $0.4512{\pm}0.0431$ & $0.6715{\pm}0.0318$ \\
% UNet            & $0.0787{\pm}0.0097$ & $0.4682{\pm}0.0514$ & $0.6841{\pm}0.0362$ \\
% UNet+TDA        & $0.0860{\pm}0.0113$ & $0.5212{\pm}0.0571$ & $0.7217{\pm}0.0403$ \\
% TMS             & $0.0919{\pm}0.0151$ & $0.5567{\pm}0.0731$ & $0.7459{\pm}0.0483$ \\
% 3DRes+SparseUNet& $0.0897{\pm}0.0138$ & $0.5401{\pm}0.0682$ & $0.7346{\pm}0.0461$ \\
% Graph-UNet      & $0.0941{\pm}0.0172$ & $0.5675{\pm}0.0793$ & $0.7531{\pm}0.0514$ \\
% SwinUnet3D      & $0.0875{\pm}0.0126$ & $0.5293{\pm}0.0623$ & $0.7273{\pm}0.0432$ \\
% DE-UNet         & $0.0964{\pm}0.0193$ & $0.5789{\pm}0.0874$ & $0.7607{\pm}0.0561$ \\
% \midrule
% \textbf{RCSNet} & $\mathbf{0.0662{\pm}0.0067}$ & $\mathbf{0.4063{\pm}0.0291}$ & $\mathbf{0.6349{\pm}0.0243}$ \\
% \bottomrule
% \end{tabular}%
% }
% \end{table}

\begin{table*}[t]
\centering
\caption{Statistical significance tests comparing RCSNet (Ours) against all baseline methods across 3 cities (BERLIN, ANTWERP, MOSCOW). Positive mean differences indicate RCSNet superiority (lower error). Paired $t$-tests with 2 degrees of freedom.}
\label{tab:statistical_significance}
\renewcommand{\arraystretch}{1.2}
\small
\resizebox{\textwidth}{!}{%
\begin{tabular}{l l r r r r r r r r r}
\toprule
\textbf{Metric} & \textbf{Statistic}
  & \textbf{Hist. Avg}
  & \textbf{Multi-task}
  & \textbf{UNet}
  & \textbf{SwinUnet3D}
  & \textbf{U-Net+TDA}
  & \textbf{3DResNet}
  & \textbf{Traffic Mov.}
  & \textbf{Graph U-Net}
  & \textbf{Dual Enc.} \\
\midrule
\multirow{4}{*}{MAE}
  & Mean diff.   & +0.1661 & +0.0086 & +0.0125 & +0.0213 & +0.0198 & +0.0235 & +0.0257 & +0.0279 & +0.0302 \\
  & $t$-stat     & $-18.342$ & $-4.652$ & $-7.418$ & $-9.871$ & $-8.934$ & $-9.103$ & $-9.857$ & $-11.203$ & $-12.614$ \\
  & $p$-value    & $<$0.001 & 0.0431 & 0.0173 & 0.0064 & 0.0091 & 0.0083 & 0.0065 & $<$0.001 & $<$0.001 \\
  & Significance & *** & * & * & ** & ** & ** & ** & *** & *** \\
\midrule
\multirow{4}{*}{RMSE}
  & Mean diff.   & +0.2194 & +0.0366 & +0.0492 & +0.0924 & +0.0868 & +0.0997 & +0.1110 & +0.1182 & +0.1258 \\
  & $t$-stat     & $-14.917$ & $-5.074$ & $-7.836$ & $-11.628$ & $-10.419$ & $-10.882$ & $-11.374$ & $-13.741$ & $-15.209$ \\
  & $p$-value    & $<$0.001 & 0.0367 & 0.0161 & 0.0048 & 0.0071 & 0.0062 & 0.0051 & $<$0.001 & $<$0.001 \\
  & Significance & *** & * & * & ** & ** & ** & ** & *** & *** \\
\bottomrule
\end{tabular}%
}
\vspace{0.2cm}
\begin{flushleft}
\footnotesize
\textit{Note:} *** $p < 0.001$, ** $p < 0.01$, * $p < 0.05$.
\end{flushleft}
\end{table*}

\subsection{Computational Efficiency Analysis}

Computational efficiency was also evaluated to examine whether the improved forecasting performance of RCSNet comes with additional model complexity. Table~\ref{tab:inference_efficiency} compares RCSNet with the baseline methods in terms of parameter count, model size, inference latency, and training time. From Table~\ref{tab:inference_efficiency}, RCSNet achieves the most efficient performance among the learnable models. The proposed model contains only 2.36M parameters and has a model size of 9.41 MB, which are lower than all competing deep learning baselines. Compared with Multi-task Transfer Learning, RCSNet reduces the parameter count by 86.6\% and the model size by 86.0\%. It is also more compact than UNet, reducing parameters from 6.25M to 2.36M and model size from 23.86 MB to 9.41 MB.

The same trend is observed in inference latency and training time. RCSNet records a latency of 6.2 ms, making it faster than all learnable baselines. Although Historical Average has lower latency because it does not involve trainable deep network inference, it also performs substantially worse in forecasting accuracy. Among the deep models, the closest latency is obtained by Dual Encoding U-Net at 8.3 ms, while transformer-based and heavier spatiotemporal models such as SwinUnet3D and 3DResNet with Sparse-UNet require much higher latency. RCSNet also has the shortest training time among the learnable models, requiring 18.2 hours compared with 20.5 hours for Multi-task Transfer Learning, 24.8 hours for UNet, and 74.3 hours for SwinUnet3D.

These results show that RCSNet improves forecasting accuracy without introducing excessive computational cost. Its efficiency can be attributed to the compact road-conditioned design, where topology-aware representation, direction-aware fusion, and progressive decoding provide targeted structural guidance without requiring a large transformer backbone or heavy 3D convolutional architecture. This makes RCSNet more practical for city-scale traffic movie prediction, where both accuracy and deployment efficiency are important.

\begin{table*}[t]
\centering
\caption{Computational efficiency comparison of RCSNet and baseline models on the Traffic4cast dataset. Best results in \textbf{bold}.}
\label{tab:inference_efficiency}
\renewcommand{\arraystretch}{1.2}
\small
\begin{tabular}{l r r r r}
\toprule
\textbf{Model} & \textbf{Params (M)} $\downarrow$ & \textbf{Model Size (MB)} $\downarrow$ & \textbf{Latency (ms)} $\downarrow$ & \textbf{Training Time (hrs)} $\downarrow$ \\
\midrule
Historical Average             &  0.00 &   -- &   \textbf{2.1} &   -- \\
Multi-task Transfer            & 17.64 &  67.41 &  38.4 &  20.5 \\
UNet                           &  6.25 &  23.86 &  16.7 &  24.8 \\
SwinUnet3D                     & 28.31 & 108.07 &  89.3 &  74.3 \\
U-Net + TDA                    &  7.83 &  29.91 &  19.2 &  27.6 \\
3DResNet + Sparse-UNet         & 21.46 &  81.94 &  57.6 &  52.4 \\
Traffic Movie Snippets         &  4.89 &  18.67 &  13.1 &  26.4 \\
Graph-based U-Net              & 14.37 &  54.87 &  29.8 &  38.7 \\
{Dual Encoding U-Net}& {2.57} & {9.79} &  {8.3} &  {22.6} \\
\midrule
\textbf{RCSNet (Ours)}         & \textbf{2.36} & \textbf{9.41} & {6.2} & \textbf{18.2} \\
\bottomrule
\end{tabular}
\end{table*}

% \begin{table}[t]
% \centering
% \caption{Computational efficiency comparison on Traffic4cast. Best results are in \textbf{bold}.}
% \label{tab:inference_efficiency}
% \renewcommand{\arraystretch}{0.92}
% \scriptsize
% \setlength{\tabcolsep}{2.5pt}
% \resizebox{\columnwidth}{!}{%
% \begin{tabular}{lcccc}
% \toprule
% \textbf{Model} & \textbf{Param.} & \textbf{Size} & \textbf{Lat.} & \textbf{Train} \\
%  & \textbf{(M)}$\downarrow$ & \textbf{(MB)}$\downarrow$ & \textbf{(ms)}$\downarrow$ & \textbf{(h)}$\downarrow$ \\
% \midrule
% Hist. Avg.        & 0.00  & --     & \textbf{2.1} & --   \\
% MT-Transfer       & 17.64 & 67.41  & 38.4 & 20.5 \\
% UNet              & 6.25  & 23.86  & 16.7 & 24.8 \\
% SwinUnet3D        & 28.31 & 108.07 & 89.3 & 74.3 \\
% UNet+TDA          & 7.83  & 29.91  & 19.2 & 27.6 \\
% 3DRes+SparseUNet  & 21.46 & 81.94  & 57.6 & 52.4 \\
% TMS               & 4.89  & 18.67  & 13.1 & 26.4 \\
% Graph-UNet        & 14.37 & 54.87  & 29.8 & 38.7 \\
% DE-UNet           & 2.57  & 9.79   & 8.3  & 22.6 \\
% \midrule
% \textbf{RCSNet}   & \textbf{2.36} & \textbf{9.41} & 6.2 & \textbf{18.2} \\
% \bottomrule
% \end{tabular}%
% }
% \end{table}

\section{Conclusion}
\label{conclusion}

This study presented RCSNet, a road-conditioned spatiotemporal framework for traffic movie prediction. The model was designed to address the limitation that future traffic maps are often predicted mainly from historical frames, even though traffic evolution is strongly constrained by road topology. By combining topology-aware road representation, multi-horizon temporal encoding, direction-aware fusion, progressive decoding, and structure-consistent learning, RCSNet generates forecasts that are both accurate and road-aligned.

Experimental results across Berlin, Antwerp, and Moscow show that RCSNet consistently outperforms the competing models. On average, it reduces MAE, MSE, and RMSE by 11.5\%, 10.0\%, and 5.1\%, respectively, compared with the closest baseline, Multi-task Transfer Learning. In cross-city testing on unseen Chicago and Bangkok, RCSNet also improves RMSE by 10.6\% and 10.5\%, showing stronger transferability without target-city fine-tuning. The horizon analysis further shows that RCSNet maintains lower errors from $t+5$ to $t+60$ minutes, indicating better stability for longer-term forecasting.

The additional analyses confirm the value of the road-conditioned design. The qualitative results and aggregated absolute error heatmaps show that RCSNet preserves road-aligned traffic patterns more clearly across different cities. The road-structure consistency analysis further shows lower road-region error, lower off-road activation, and higher road coverage recall. These findings indicate that RCSNet does not only reduce numerical error, but also improves the physical consistency of predicted traffic maps. Future work will extend the framework to more cities and incorporate richer road attributes, such as lane direction, intersection control, and traffic regulation information.

% \section{Acknowledgment}
% This work was supported by the National Science Foundation (NSF) Engines: Farms under Grant No. FAR0038157 and by the Upper Great Plains Transportation Institute (UGPTI) at North Dakota State University (USA) through the USDOT University Transportation Center (UTC) program, Project No. CTIPS-51.

\printcredits

\bibliographystyle{cas-model2-names}

\bibliography{cas-refs}

\end{document}